%% file: main.tex
\documentclass{article}

\usepackage[final]{neurips_2025}
\usepackage{subcaption}

\usepackage{graphicx}
\input{packages}

\title{Improved Training Technique for Shortcut Models}

\author{%
  Anh Nguyen\thanks{Equal contribution}\hspace{1.15em}
  Viet Nguyen\footnotemark[1]\hspace{0.4em}\thanks{Work done while at Qualcomm.}\hspace{1.15em}
  Duc Vu\hspace{1.15em}
  Trung Dao\hspace{1.15em}
  Chi Tran\hspace{1.15em}
  Toan Tran\hspace{1.15em}
  Anh Tran \\\\
  Qualcomm AI Research\thanks{Qualcomm AI Research is an initiative of Qualcomm Technologies, Inc.}
}

\begin{document}
\input{commands}

\maketitle

 \begin{abstract}
  Shortcut models represent a promising, non-adversarial paradigm for generative modeling, uniquely supporting one-step, few-step, and multi-step sampling from a single trained network. However, their widespread adoption has been stymied by critical performance bottlenecks. This paper tackles the five core issues that held shortcut models back: (1) the hidden \textbf{flaw of compounding guidance}, which we are the first to formalize, causing severe image artifacts; (2) \textbf{inflexible fixed guidance} that restricts inference-time control; (3) a pervasive \textbf{frequency bias} driven by a reliance on low-level distances in the direct domain, which biases reconstructions toward low frequencies; (4) \textbf{divergent self-consistency} arising from a conflict with EMA training; and (5) \textbf{curvy flow trajectories} that impede convergence. To address these challenges, we introduce iSM, a unified training framework that systematically resolves each limitation. Our framework is built on four key improvements: \textit{Intrinsic Guidance} provides explicit, dynamic control over guidance strength, resolving both compounding guidance and inflexibility. A \textit{Multi-Level Wavelet Loss} mitigates frequency bias to restore high-frequency details. \textit{Scaling Optimal Transport (sOT)} reduces training variance and learns straighter, more stable generative paths. Finally, a \textit{Twin EMA} strategy reconciles training stability with self-consistency. Extensive experiments on ImageNet $256\times256$ demonstrate that our approach yields substantial FID improvements over baseline shortcut models across one-step, few-step, and multi-step generation, making shortcut models a viable and competitive class of generative models.
\end{abstract}

\input{tex/introduction}
\input{tex/background}

\input{tex/method}
\input{tex/experiment}
\input{tex/conclusion}

\newpage
\bibliography{references}
\bibliographystyle{abbrv}
\newpage
\input{checklist}

\newpage
\appendix

\input{tex/appendix}




\end{document}

%% file: packages.tex
\usepackage[utf8]{inputenc} 
\usepackage[T1]{fontenc}    
\usepackage[english]{babel} 
\usepackage{afterpage} 

\usepackage{hyperref}       
\usepackage{url}            
\usepackage{microtype}      
\usepackage{xspace}         
\usepackage{soul}           
\usepackage{mdframed}       
\usepackage{comment}        
\usepackage{bm}             

\usepackage{mathtools}
\usepackage{amsthm}
\usepackage{amsfonts}       
\usepackage{amssymb}
\usepackage{mathtools}      
\usepackage{nicefrac}       
\usepackage{latexsym}       
\usepackage{stackrel}       
\usepackage{cases}          
\usepackage{stmaryrd}       
\usepackage{mathabx}        
\usepackage{dsfont}         

\usepackage{booktabs}       
\usepackage{array}          
\usepackage{multirow}       
\usepackage{colortbl}       
\usepackage{makecell}       
\usepackage{diagbox}        
\usepackage{longtable}      
\usepackage{tabularx}       
\usepackage{bigstrut}       
\usepackage{arydshln}       

\usepackage{graphicx}       
\usepackage{float}          
\usepackage[font=small,skip=0pt]{caption} 
\usepackage{wrapfig,lipsum,booktabs}
\usepackage{tikz}           
\usepackage{subcaption}     
\usepackage{sidecap}        
\usepackage[export]{adjustbox} 

\usepackage{thm-restate}    
\usepackage[capitalise]{cleveref} 
\usepackage{theoremref}     

\usepackage{xcolor}         
\usepackage{xkcdcolors}     
\usepackage{fancyvrb}       
\usepackage[most]{tcolorbox}

\usepackage{algorithm}
\usepackage{algpseudocode}
\usepackage{listings}
\usepackage{caption}

\definecolor{deepgreen}{RGB}{0,200,0}

%% file: commands.tex
\definecolor{burntorange}{RGB}{204, 85, 0}      
\definecolor{crimson}{RGB}{220, 20, 60}        
\definecolor{teal}{RGB}{0, 128, 128}           
\definecolor{indigo}{RGB}{75, 0, 130}          
\definecolor{royalblue}{RGB}{65, 105, 225}     
\definecolor{magenta}{RGB}{255, 0, 255}        

\newcommand{\anh}[1]{\textcolor{orange}{[Anh Tran: #1]}}
\newcommand{\anhn}[1]{\textcolor{magenta}{[Anh Nguyen: #1]}}
\newcommand{\viet}[1]{\textcolor{teal}{[Viet: #1]}}
\newcommand{\ducvh}[1]{\textcolor{red}{[Duc: #1]}}
\newcommand{\toan}[1]{\textcolor{magenta}{[Toan: #1]}}
\newcommand{\chitran}[1]{\textcolor{crimson}{[Chi: #1]}}
\newcommand{\trung}[1]{\textcolor{purple}{[Trung: #1]}}

\newcommand{\R}{\mathbb{R}} 
\newcommand{\N}{\mathcal{N}} 
\newcommand{\Ls}{\mathcal{L}} 
\newcommand{\velocity}{\mathrm{velocity}} 
\newcommand{\guidance}{\mathrm{guidance}} 
\newcommand{\consistency}{\mathrm{consistency}} 
\newcommand{\interval}{\mathrm{interval}} 
\newcommand{\CFG}{\mathrm{CFG}} 
\newcommand{\LL}{\mathrm{LL}} 
\newcommand{\HL}{\mathrm{HL}} 
\newcommand{\LH}{\mathrm{LH}} 
\newcommand{\HH}{\mathrm{HH}} 
\newcommand{\wavelet}{\mathrm{wavelet}} 
\newcommand{\total}{\mathrm{total}} 
\newcommand{\DWT}{\mathrm{DWT}} 
\newcommand{\pred}{\mathrm{pred}} 
\newcommand{\target}{\mathrm{target}} 
\newcommand{\twin}{\mathrm{twin}} 
\newcommand{\infer}{\mathrm{infer}} 
\newcommand{\K}{\mathrm{K}} 
\newcommand{\M}{\mathrm{M}} 
\newcommand{\wmax}{\mathrm{max}} 
\newcommand{\one}{N=1} 
\newcommand{\four}{N=4} 

\newcommand{\varnothingcond}{\varnothing} 
\newcommand{\xt}{\bm{x}_t} 
\newcommand{\x}{\bm{x}} 
\newcommand{\xzero}{\bm{x}_0} 
\newcommand{\xone}{\bm{x}_1} 
\newcommand{\gtheta}{\bm{g}_\theta} 
\newcommand{\stheta}{\bm{s}_\theta} 
\newcommand{\sthetaema}{\bm{s}_\theta^{-}} 
\newcommand{\sv}{\bm{s}_{\velocity}} 
\newcommand{\sg}{\bm{s}_{\guidance}} 
\newcommand{\scon}{\bm{s}_{\consistency}} 
\newcommand{\stopgrad}[1]{\mathrm{sg}(#1)} 

\def\rvtheta{{\bm{\theta}}}
\newcommand{\abs}[1]{\lvert#1\rvert}
\newcommand*{\tran}{^{\mkern-1.5mu\mathsf{T}}}
\newcommand{\mbf}[1]{\mathbf{#1}}
\newcommand{\mbb}[1]{\mathbb{#1}}
\newcommand{\mcal}[1]{\mathcal{#1}}
\newcommand{\mrel}{\mathrel}

\def\eg{\emph{e.g}\onedot}
\def\Eg{\emph{E.g}\onedot}
\def\ie{\emph{i.e}\onedot}
\def\viz{\emph{viz}\onedot}
\def\Ie{\emph{I.e}\onedot}
\def\Viz{\emph{Viz}\onedot}
\def\cf{\emph{cf}\onedot}
\def\Cf{\emph{Cf}\onedot}
\def\etc{\emph{etc}\onedot}
\def\vs{\emph{vs}\onedot}
\def\wrt{w.r.t\onedot}
\def\resp{resp\onedot}
\def\dof{d.o.f\onedot}
\def\aka{a.k.a\onedot}
\def\iid{i.i.d\onedot}
\def\etal{\emph{et al}\onedot}

\definecolor{darkgreen}{rgb}{0,0.6,0}
\newcommand{\propositionbox}[1]{
    \begin{tcolorbox}[
        colback=white!90!gray,     
        colframe=teal!60!black,     
        arc=5pt,                    
        boxsep=5pt,                 
        left=10pt,                  
        right=10pt,                 
        top=2pt,                    
        bottom=2pt,                 
        boxrule=0.8pt,              
        drop shadow=gray!50!white,  
        enhanced jigsaw             
    ]
    \vspace{-0.1cm}
        #1
    \end{tcolorbox}
}

\newtheorem{observation}{Observation}
\newtheorem{hypothesis}{Hypothesis}
\newtheorem{claim}{Claim}
\newtheorem{theorem}{Theorem}
\newtheorem{definition}{Definition}
\newtheorem{assumption}{Assumption}
\newtheorem{proposition}{Proposition}
\newtheorem{corollary}{Corollary}
\newtheorem{lemma}{Lemma}
\newtheorem{remark}{Remark}
\newcommand{\figleft}{{\em (Left)}}
\newcommand{\figcenter}{{\em (Center)}}
\newcommand{\figright}{{\em (Right)}}
\newcommand{\figtop}{{\em (Top)}}
\newcommand{\figbottom}{{\em (Bottom)}}
\newcommand{\captiona}{{\em (a)}}
\newcommand{\captionb}{{\em (b)}}
\newcommand{\captionc}{{\em (c)}}
\newcommand{\captiond}{{\em (d)}}

\newcommand{\newterm}[1]{{\bf #1}}

\def\figref#1{figure~\ref{#1}}
\def\Figref#1{Figure~\ref{#1}}
\def\twofigref#1#2{figures \ref{#1} and \ref{#2}}
\def\quadfigref#1#2#3#4{figures \ref{#1}, \ref{#2}, \ref{#3} and \ref{#4}}
\def\secref#1{section~\ref{#1}}
\def\Secref#1{Section~\ref{#1}}
\def\twosecrefs#1#2{sections \ref{#1} and \ref{#2}}
\def\secrefs#1#2#3{sections \ref{#1}, \ref{#2} and \ref{#3}}
\def\eqref#1{equation~\ref{#1}}
\def\Eqref#1{Equation~\ref{#1}}
\def\plaineqref#1{\ref{#1}}
\def\chapref#1{chapter~\ref{#1}}
\def\Chapref#1{Chapter~\ref{#1}}
\def\rangechapref#1#2{chapters\ref{#1}--\ref{#2}}
\def\algref#1{algorithm~\ref{#1}}
\def\Algref#1{Algorithm~\ref{#1}}
\def\twoalgref#1#2{algorithms \ref{#1} and \ref{#2}}
\def\Twoalgref#1#2{Algorithms \ref{#1} and \ref{#2}}
\def\partref#1{part~\ref{#1}}
\def\Partref#1{Part~\ref{#1}}
\def\twopartref#1#2{parts \ref{#1} and \ref{#2}}

\def\ceil#1{\lceil #1 \rceil}
\def\floor#1{\lfloor #1 \rfloor}
\def\1{\bm{1}}
\newcommand{\train}{\mathcal{D}}
\newcommand{\valid}{\mathcal{D_{\mathrm{valid}}}}
\newcommand{\test}{\mathcal{D_{\mathrm{test}}}}

\def\eps{{\epsilon}}

\def\reta{{\textnormal{$\eta$}}}
\def\ra{{\textnormal{a}}}
\def\rb{{\textnormal{b}}}
\def\rc{{\textnormal{c}}}
\def\rd{{\textnormal{d}}}
\def\re{{\textnormal{e}}}
\def\rf{{\textnormal{f}}}
\def\rg{{\textnormal{g}}}
\def\rh{{\textnormal{h}}}
\def\ri{{\textnormal{i}}}
\def\rj{{\textnormal{j}}}
\def\rk{{\textnormal{k}}}
\def\rl{{\textnormal{l}}}
\def\rn{{\textnormal{n}}}
\def\ro{{\textnormal{o}}}
\def\rp{{\textnormal{p}}}
\def\rq{{\textnormal{q}}}
\def\rr{{\textnormal{r}}}
\def\rs{{\textnormal{s}}}
\def\rt{{\textnormal{t}}}
\def\ru{{\textnormal{u}}}
\def\rv{{\textnormal{v}}}
\def\rw{{\textnormal{w}}}
\def\rx{{\textnormal{x}}}
\def\ry{{\textnormal{y}}}
\def\rz{{\textnormal{z}}}

\def\rvepsilon{{\bm{\epsilon}}}
\def\rvtheta{{\bm{\theta}}}
\def\rva{{\mathbf{a}}}
\def\rvb{{\mathbf{b}}}
\def\rvc{{\mathbf{c}}}
\def\rvd{{\mathbf{d}}}
\def\rve{{\mathbf{e}}}
\def\rvf{{\mathbf{f}}}
\def\rvg{{\mathbf{g}}}
\def\rvh{{\mathbf{h}}}
\def\rvu{{\mathbf{i}}}
\def\rvj{{\mathbf{j}}}
\def\rvk{{\mathbf{k}}}
\def\rvl{{\mathbf{l}}}
\def\rvm{{\mathbf{m}}}
\def\rvn{{\mathbf{n}}}
\def\rvo{{\mathbf{o}}}
\def\rvp{{\mathbf{p}}}
\def\rvq{{\mathbf{q}}}
\def\rvr{{\mathbf{r}}}
\def\rvs{{\mathbf{s}}}
\def\rvt{{\mathbf{t}}}
\def\rvu{{\mathbf{u}}}
\def\rvv{{\mathbf{v}}}
\def\rvw{{\mathbf{w}}}
\def\rvx{{\mathbf{x}}}
\def\rvy{{\mathbf{y}}}
\def\rvz{{\mathbf{z}}}

\def\erva{{\textnormal{a}}}
\def\ervb{{\textnormal{b}}}
\def\ervc{{\textnormal{c}}}
\def\ervd{{\textnormal{d}}}
\def\erve{{\textnormal{e}}}
\def\ervf{{\textnormal{f}}}
\def\ervg{{\textnormal{g}}}
\def\ervh{{\textnormal{h}}}
\def\ervi{{\textnormal{i}}}
\def\ervj{{\textnormal{j}}}
\def\ervk{{\textnormal{k}}}
\def\ervl{{\textnormal{l}}}
\def\ervm{{\textnormal{m}}}
\def\ervn{{\textnormal{n}}}
\def\ervo{{\textnormal{o}}}
\def\ervp{{\textnormal{p}}}
\def\ervq{{\textnormal{q}}}
\def\ervr{{\textnormal{r}}}
\def\ervs{{\textnormal{s}}}
\def\ervt{{\textnormal{t}}}
\def\ervu{{\textnormal{u}}}
\def\ervv{{\textnormal{v}}}
\def\ervw{{\textnormal{w}}}
\def\ervx{{\textnormal{x}}}
\def\ervy{{\textnormal{y}}}
\def\ervz{{\textnormal{z}}}

\def\rmA{{\mathbf{A}}}
\def\rmB{{\mathbf{B}}}
\def\rmC{{\mathbf{C}}}
\def\rmD{{\mathbf{D}}}
\def\rmE{{\mathbf{E}}}
\def\rmF{{\mathbf{F}}}
\def\rmG{{\mathbf{G}}}
\def\rmH{{\mathbf{H}}}
\def\rmI{{\mathbf{I}}}
\def\rmJ{{\mathbf{J}}}
\def\rmK{{\mathbf{K}}}
\def\rmL{{\mathbf{L}}}
\def\rmM{{\mathbf{M}}}
\def\rmN{{\mathbf{N}}}
\def\rmO{{\mathbf{O}}}
\def\rmP{{\mathbf{P}}}
\def\rmQ{{\mathbf{Q}}}
\def\rmR{{\mathbf{R}}}
\def\rmS{{\mathbf{S}}}
\def\rmT{{\mathbf{T}}}
\def\rmU{{\mathbf{U}}}
\def\rmV{{\mathbf{V}}}
\def\rmW{{\mathbf{W}}}
\def\rmX{{\mathbf{X}}}
\def\rmY{{\mathbf{Y}}}
\def\rmZ{{\mathbf{Z}}}

\def\ermA{{\textnormal{A}}}
\def\ermB{{\textnormal{B}}}
\def\ermC{{\textnormal{C}}}
\def\ermD{{\textnormal{D}}}
\def\ermE{{\textnormal{E}}}
\def\ermF{{\textnormal{F}}}
\def\ermG{{\textnormal{G}}}
\def\ermH{{\textnormal{H}}}
\def\ermI{{\textnormal{I}}}
\def\ermJ{{\textnormal{J}}}
\def\ermK{{\textnormal{K}}}
\def\ermL{{\textnormal{L}}}
\def\ermM{{\textnormal{M}}}
\def\ermN{{\textnormal{N}}}
\def\ermO{{\textnormal{O}}}
\def\ermP{{\textnormal{P}}}
\def\ermQ{{\textnormal{Q}}}
\def\ermR{{\textnormal{R}}}
\def\ermS{{\textnormal{S}}}
\def\ermT{{\textnormal{T}}}
\def\ermU{{\textnormal{U}}}
\def\ermV{{\textnormal{V}}}
\def\ermW{{\textnormal{W}}}
\def\ermX{{\textnormal{X}}}
\def\ermY{{\textnormal{Y}}}
\def\ermZ{{\textnormal{Z}}}

\def\vzero{{\bm{0}}}
\def\vone{{\bm{1}}}
\def\vmu{{\bm{\mu}}}
\def\vtheta{{\bm{\theta}}}
\def\valpha{{\bm{\alpha}}}
\def\vepsilon{{\bm{\epsilon}}}
\def\vpi{{\bm{\pi}}}
\def\vomega{{\bm{\omega}}}
\def\vlambda{{\bm{\lambda}}}
\def\vphi{{\bm{\phi}}}
\def\va{{\bm{a}}}
\def\vb{{\bm{b}}}
\def\vc{{\bm{c}}}
\def\vd{{\bm{d}}}
\def\ve{{\bm{e}}}
\def\vf{{\bm{f}}}
\def\vg{{\bm{g}}}
\def\vh{{\bm{h}}}
\def\vi{{\bm{i}}}
\def\vj{{\bm{j}}}
\def\vk{{\bm{k}}}
\def\vl{{\bm{l}}}
\def\vm{{\bm{m}}}
\def\vn{{\bm{n}}}
\def\vo{{\bm{o}}}
\def\vp{{\bm{p}}}
\def\vq{{\bm{q}}}
\def\vr{{\bm{r}}}
\def\vs{{\bm{s}}}
\def\vt{{\bm{t}}}
\def\vu{{\bm{u}}}
\def\vv{{\bm{v}}}
\def\vw{{\bm{w}}}
\def\vx{{\bm{x}}}
\def\vy{{\bm{y}}}
\def\vz{{\bm{z}}}

\def\evalpha{{\alpha}}
\def\evbeta{{\beta}}
\def\evepsilon{{\epsilon}}
\def\evlambda{{\lambda}}
\def\evomega{{\omega}}
\def\evmu{{\mu}}
\def\evpsi{{\psi}}
\def\evsigma{{\sigma}}
\def\evtheta{{\theta}}
\def\eva{{a}}
\def\evb{{b}}
\def\evc{{c}}
\def\evd{{d}}
\def\eve{{e}}
\def\evf{{f}}
\def\evg{{g}}
\def\evh{{h}}
\def\evi{{i}}
\def\evj{{j}}
\def\evk{{k}}
\def\evl{{l}}
\def\evm{{m}}
\def\evn{{n}}
\def\evo{{o}}
\def\evp{{p}}
\def\evq{{q}}
\def\evr{{r}}
\def\evs{{s}}
\def\evt{{t}}
\def\evu{{u}}
\def\evv{{v}}
\def\evw{{w}}
\def\evx{{x}}
\def\evy{{y}}
\def\evz{{z}}

\def\mA{{\bm{A}}}
\def\mB{{\bm{B}}}
\def\mC{{\bm{C}}}
\def\mD{{\bm{D}}}
\def\mE{{\bm{E}}}
\def\mF{{\bm{F}}}
\def\mG{{\bm{G}}}
\def\mH{{\bm{H}}}
\def\mI{{\bm{I}}}
\def\mJ{{\bm{J}}}
\def\mK{{\bm{K}}}
\def\mL{{\bm{L}}}
\def\mM{{\bm{M}}}
\def\mN{{\bm{N}}}
\def\mO{{\bm{O}}}
\def\mP{{\bm{P}}}
\def\mQ{{\bm{Q}}}
\def\mR{{\bm{R}}}
\def\mS{{\bm{S}}}
\def\mT{{\bm{T}}}
\def\mU{{\bm{U}}}
\def\mV{{\bm{V}}}
\def\mW{{\bm{W}}}
\def\mX{{\bm{X}}}
\def\mY{{\bm{Y}}}
\def\mZ{{\bm{Z}}}
\def\mBeta{{\bm{\beta}}}
\def\mPhi{{\bm{\Phi}}}
\def\mLambda{{\bm{\Lambda}}}
\def\mSigma{{\bm{\Sigma}}}

\newcommand{\tens}[1]{\bm{\mathsfit{#1}}}
\def\tA{{\tens{A}}}
\def\tB{{\tens{B}}}
\def\tC{{\tens{C}}}
\def\tD{{\tens{D}}}
\def\tE{{\tens{E}}}
\def\tF{{\tens{F}}}
\def\tG{{\tens{G}}}
\def\tH{{\tens{H}}}
\def\tI{{\tens{I}}}
\def\tJ{{\tens{J}}}
\def\tK{{\tens{K}}}
\def\tL{{\tens{L}}}
\def\tM{{\tens{M}}}
\def\tN{{\tens{N}}}
\def\tO{{\tens{O}}}
\def\tP{{\tens{P}}}
\def\tQ{{\tens{Q}}}
\def\tR{{\tens{R}}}
\def\tS{{\tens{S}}}
\def\tT{{\tens{T}}}
\def\tU{{\tens{U}}}
\def\tV{{\tens{V}}}
\def\tW{{\tens{W}}}
\def\tX{{\tens{X}}}
\def\tY{{\tens{Y}}}
\def\tZ{{\tens{Z}}}

\def\gA{{\mathcal{A}}}
\def\gB{{\mathcal{B}}}
\def\gC{{\mathcal{C}}}
\def\gD{{\mathcal{D}}}
\def\gE{{\mathcal{E}}}
\def\gF{{\mathcal{F}}}
\def\gG{{\mathcal{G}}}
\def\gH{{\mathcal{H}}}
\def\gI{{\mathcal{I}}}
\def\gJ{{\mathcal{J}}}
\def\gK{{\mathcal{K}}}
\def\gL{{\mathcal{L}}}
\def\gM{{\mathcal{M}}}
\def\gN{{\mathcal{N}}}
\def\gO{{\mathcal{O}}}
\def\gP{{\mathcal{P}}}
\def\gQ{{\mathcal{Q}}}
\def\gR{{\mathcal{R}}}
\def\gS{{\mathcal{S}}}
\def\gT{{\mathcal{T}}}
\def\gU{{\mathcal{U}}}
\def\gV{{\mathcal{V}}}
\def\gW{{\mathcal{W}}}
\def\gX{{\mathcal{X}}}
\def\gY{{\mathcal{Y}}}
\def\gZ{{\mathcal{Z}}}

\def\sA{{\mathbb{A}}}
\def\sB{{\mathbb{B}}}
\def\sC{{\mathbb{C}}}
\def\sD{{\mathbb{D}}}
\def\sF{{\mathbb{F}}}
\def\sG{{\mathbb{G}}}
\def\sH{{\mathbb{H}}}
\def\sI{{\mathbb{I}}}
\def\sJ{{\mathbb{J}}}
\def\sK{{\mathbb{K}}}
\def\sL{{\mathbb{L}}}
\def\sM{{\mathbb{M}}}
\def\sN{{\mathbb{N}}}
\def\sO{{\mathbb{O}}}
\def\sP{{\mathbb{P}}}
\def\sQ{{\mathbb{Q}}}
\def\sR{{\mathbb{R}}}
\def\sS{{\mathbb{S}}}
\def\sT{{\mathbb{T}}}
\def\sU{{\mathbb{U}}}
\def\sV{{\mathbb{V}}}
\def\sW{{\mathbb{W}}}
\def\sX{{\mathbb{X}}}
\def\sY{{\mathbb{Y}}}
\def\sZ{{\mathbb{Z}}}

\def\emLambda{{\Lambda}}
\def\emA{{A}}
\def\emB{{B}}
\def\emC{{C}}
\def\emD{{D}}
\def\emE{{E}}
\def\emF{{F}}
\def\emG{{G}}
\def\emH{{H}}
\def\emI{{I}}
\def\emJ{{J}}
\def\emK{{K}}
\def\emL{{L}}
\def\emM{{M}}
\def\emN{{N}}
\def\emO{{O}}
\def\emP{{P}}
\def\emQ{{Q}}
\def\emR{{R}}
\def\emS{{S}}
\def\emT{{T}}
\def\emU{{U}}
\def\emV{{V}}
\def\emW{{W}}
\def\emX{{X}}
\def\emY{{Y}}
\def\emZ{{Z}}
\def\emSigma{{\Sigma}}

\newcommand{\etens}[1]{\mathsfit{#1}}
\def\iidsim{{\stackrel{\text{\iid}}{\sim}}}
\def\dvert{{~\|~}}
\def\etLambda{{\etens{\Lambda}}}
\def\etA{{\etens{A}}}
\def\etB{{\etens{B}}}
\def\etC{{\etens{C}}}
\def\etD{{\etens{D}}}
\def\etE{{\etens{E}}}
\def\etF{{\etens{F}}}
\def\etG{{\etens{G}}}
\def\etH{{\etens{H}}}
\def\etI{{\etens{I}}}
\def\etJ{{\etens{J}}}
\def\etK{{\etens{K}}}
\def\etL{{\etens{L}}}
\def\etM{{\etens{M}}}
\def\etN{{\etens{N}}}
\def\etO{{\etens{O}}}
\def\etP{{\etens{P}}}
\def\etQ{{\etens{Q}}}
\def\etR{{\etens{R}}}
\def\etS{{\etens{S}}}
\def\etT{{\etens{T}}}
\def\etU{{\etens{U}}}
\def\etV{{\etens{V}}}
\def\etW{{\etens{W}}}
\def\etX{{\etens{X}}}
\def\etY{{\etens{Y}}}
\def\etZ{{\etens{Z}}}

\newcommand{\pdata}{p_{\rm{data}}}
\newcommand{\ptrain}{\hat{p}_{\rm{data}}}
\newcommand{\Ptrain}{\hat{P}_{\rm{data}}}
\newcommand{\pmodel}{p_{\rm{model}}}
\newcommand{\Pmodel}{P_{\rm{model}}}
\newcommand{\ptildemodel}{\tilde{p}_{\rm{model}}}
\newcommand{\pencode}{p_{\rm{encoder}}}
\newcommand{\pdecode}{p_{\rm{decoder}}}
\newcommand{\precons}{p_{\rm{reconstruct}}}

\newcommand{\laplace}{\mathrm{Laplace}} %

\newcommand{\E}{\mathbb{E}}
\newcommand{\emp}{\tilde{p}}
\newcommand{\lr}{\alpha}
\newcommand{\reg}{\lambda}
\newcommand{\rect}{\mathrm{rectifier}}
\newcommand{\softmax}{\mathrm{softmax}}
\newcommand{\sigmoid}{\sigma}
\newcommand{\softplus}{\zeta}
\newcommand{\KL}{D_{\mathrm{KL}}}
\newcommand{\Var}{\mathrm{Var}}
\newcommand{\standarderror}{\mathrm{SE}}
\newcommand{\Cov}{\mathrm{Cov}}
\newcommand{\psde}{p_\vtheta^{\textnormal{SDE}}}
\newcommand{\ptsde}{\tilde{p}_\vtheta^{\textnormal{SDE}}}
\newcommand{\pode}{p_\vtheta^{\textnormal{ODE}}}
\newcommand{\ptode}{\tilde{p}_\vtheta^{\textnormal{ODE}}}
\newcommand{\boundsm}{\mcal{L}^{\text{SM}}_\vtheta}
\newcommand{\bounddsm}{\mcal{L}^{\text{DSM}}_\vtheta}
\newcommand{\jsm}{\mcal{J}_{\textnormal{SM}}}
\newcommand{\jdsm}{\mcal{J}_{\textnormal{DSM}}}
\newcommand{\normlzero}{L^0}
\newcommand{\normlone}{L^1}
\newcommand{\normltwo}{L^2}
\newcommand{\normlp}{L^p}
\newcommand{\normmax}{L^\infty}

\colorlet{yellow}{green!10!orange}

\newcommand{\parents}{Pa} %

\newcommand{\ud}{\mathop{}\!\mathrm{d}}

\newcommand{\norm}[1]{\left\lVert#1\right\rVert}

\newcommand{\up}{\mathrm}
\def\dbar{\mathrm{\mathchar'26\mkern-12mu d}}

\definecolor{darkgreen}{rgb}{0.0, 0.5, 0.0}

\let\ab\allowbreak

\newtheorem{innercustomthm}{Theorem}
\newenvironment{customthm}[1]
{\renewcommand\theinnercustomthm{#1}\innercustomthm}
{\endinnercustomthm}
\newtheorem{innercustomlem}{Lemma}
\newenvironment{customlem}[1]
{\renewcommand\theinnercustomlem{#1}\innercustomlem}
{\endinnercustomlem}
\newtheorem{innercustomhyp}{Hypothesis}
\newenvironment{customhyp}[1]
{\renewcommand\theinnercustomhyp{#1}\innercustomhyp}
{\endinnercustomhyp}
\newtheorem{innercustomprop}{Proposition}
\newenvironment{customprop}[1]
{\renewcommand\theinnercustomprop{#1}\innercustomprop}
{\endinnercustomprop}
\newtheorem{innercustomass}{Assumption}
\newenvironment{customass}[1]
{\renewcommand\theinnercustomass{#1}\innercustomass}
{\endinnercustomass}

\newcommand{\ttimes}{{\mkern-1mu\times\mkern-1mu}}

%% file: tex/introduction.tex
\section{Introduction} \label{sec:intro}
Recent generative diffusion models based on flow matching have achieved remarkable success in synthesizing high-fidelity data across various domains \cite{sd3,flux,wan,oral}. While demonstrably successful, a significant bottleneck persists: generating samples requires integrating the learned vector field over many discrete timesteps, resulting in high computational costs that limit deployment in resource-constrained environments and latency-sensitive applications. Accelerating sampling without sacrificing quality has thus become a critical research area.

Various acceleration techniques have emerged, including timestep distillation \cite{swiftbrush,swiftbrushv2,dmd,dmd2,sanasprint,snoopi,selfcorrected}, advanced numerical solvers \cite{dpmsolver}, lightweight architectures \cite{ditair,sana,fasterdit}, and single-stage training procedures \cite{song2023consistency,ict,imm,shortcut}. Shortcut models (SM) \cite{shortcut} offer a particularly elegant approach, using a generator conditioned on both noise level $t$ and desired step size $d$ with an additional self-consistency loss. This enables the prediction of multiple timesteps ahead in a single forward pass, supporting variable sampling budgets at inference using the same network. Despite this promising method, widespread adoption has been hindered by critical performance bottlenecks. This paper tackles the five core issues that held shortcut models back: \textit{the hidden flaw of compounding guidance, inflexible fixed guidance, frequency bias, divergent self-consistency, and curvy flow trajectories}. 

The first two issues stem from a flawed integration of Classifier-Free Guidance (CFG). Not only are users locked into a fixed guidance scale at training time—sacrificing control over the diversity-fidelity trade-off—but this design also conceals a deeper flaw. We are the first to formalize that this fixed guidance \textit{compounds exponentially} across the implicit sub-steps of a large generation step, causing severe image artifacts. The other three issues further degrade sample quality: a reliance on pixel-wise losses creates a \textit{frequency bias} toward blurry images; random noise-data pairings create unstable, \textit{high-curvature generative paths}; and a \textit{temporal lag in the EMA target} prevents the model from learning true self-consistency for large jumps.

To address these challenges, we introduce the \textbf{Improved Shortcut Model (iSM)}, a unified training framework that systematically resolves each limitation through four key components.
\begin{enumerate}
    \item \textbf{Intrinsic Guidance}: We resolve both guidance-related flaws by making the guidance scale an explicit input to the model. This provides \textit{dynamic, inference-time control}, works out-of-the-box for one-step generation, and reduces inference time by $\sim$50\% compared to standard CFG, all while correcting the compounding effect.
    \item \textbf{Multi-Level Wavelet Loss}: We replace standard pixel-wise objectives with a \textit{frequency-aware loss} that forces the model to reconstruct high-frequency details, mitigating frequency bias.
    \item \textbf{Scaling Optimal Transport (sOT)}: We straighten generative trajectories by periodically pooling samples from several mini-batches to compute a large-scale transport plan. This \textit{decouples the OT batch size from the training batch size}, yielding more stable paths without a heavy computational cost.
    \item \textbf{Twin EMA Strategy}: We eliminate target lag by \textit{maintaining two EMA networks}—a fast-decay one to generate fresh, up-to-date consistency targets and a slow-decay one to ensure stable, high-quality inference.
\end{enumerate}

With these improvements, iSM achieves FID scores of 5.27 and 2.05 on ImageNet $256\times256$ with just one and four sampling steps, respectively. Our work \textit{closes the performance gap} with leading generative models and establishes improved shortcut models as a \textit{flexible, efficient, and highly competitive} modeling paradigm.

%% file: tex/background.tex
\section{Preliminaries} \label{sec:background}

\textbf{Flow Matching (FM)} \cite{lipman2022flow, liu2022flow} offers an elegant framework for generative modeling of data distributions $p_{\text{data}}(\bm{x})$. At its core, FM defines $\bm{x}_t = (1-t)\bm{x}_0 + t\bm{x}_1$ as a linear interpolation between a noise sample $\bm{x}_0$ drawn from a standard normal distribution $\mathcal{N}(0, \mathbf{I})$, denoted $\mathcal{N}$, and a data sample $\bm{x}_1$ drawn from the data distribution $D$. Here, $t \in [0, 1]$ represents the timestep, parameterizing the interpolation from noise ($t=0$) to data ($t=1$). It then trains a velocity model $\bar{\bm{v}}_\theta(\bm{x}_t, t, c)$ to match the ground-truth velocity field $\bm{v} = \xone - \xzero$ by minimizing the following velocity loss:
\begin{equation}
\mathcal{L}_\velocity(\theta) = \mathbb{E}_{\substack{\xzero \sim \N,\, (\xone, c) \sim D \\ t \sim p(t)}} [||\bar{\bm{v}}_\theta(\bm{x}_t, t, c) - \bm{v}||^2].
\end{equation}
Here, $\bm{x}_t$ is the sample interpolated at time $t$, and $c$ represents any conditioning information, such as text or class labels, associated with the data sample $\bm{x}_1$.
Sampling from such a model typically involves discretizing the learned ODE using numerical methods like Euler integration, often requiring dozens or hundreds of steps. 
Naively taking large steps with $\bar{\bm{v}}_\theta$ leads to significant discretization errors, as the predicted velocity points towards an average of potential target data points, causing mode collapse. \\
\textbf{Shortcut Models (SM).}  To address this limitation and enable efficient few-step and one-step generation, \cite{shortcut} proposes training a shortcut model that conditions the neural network $\stheta(\xt, t, c, d)$ not only on the current timestep $t$ and the condition $c$, but also on a desired step size $d$. 
This allows the model to predict the normalized displacement needed to directly reach the next point $\bm{x}'_{t+d}$ at time $t+d$, thereby bypassing intermediate steps of the probability flow ODE. 

The training approach includes \textit{flow-matching} for infinitesimal steps and \textit{self-consistency} for larger steps. At infinitesimal step sizes ($d \approx 0$), shortcut models use the flow-matching objective, regressing the model $\stheta(\xt, t, c, d=0)$ to predict the empirical velocity $\sv$, similar to traditional flow-matching models. For larger step sizes ($d>0$), shortcut models leverage a self-consistency property where one large shortcut step is equivalent to two consecutive smaller shortcut steps of half the size. This property allows the model to learn efficient large-step transitions by recursively breaking them into smaller steps.
To implement this self-consistency objective for training, they select a base number of steps $N$ ($N=128$ in practice), which defines the smallest time unit for the ODE approximation. This results in $\log_2(N)+1 = 8$ distinct shortcut lengths available during training, specifically $d \in \{1/128, 1/64, \dots, 1/2, 1\}$, over which the self-consistency loss is applied. When $d$ is at the smallest value (e.g., 1/128), they instead query the model at $d = 0$. CFG \cite{cfg}, widely used to improve conditional sample quality, is integrated within SM to form guided targets $\gtheta^w$ based on the model output $\stheta$:
\begin{align}
\gtheta^w(\xt, t, c, d) &= \stheta(\xt, t, c, d) + w \cdot (\stheta(\xt, t, c, d) - \stheta(\xt, t, \varnothingcond, d)),
\label{eq:cfg}
\end{align}
where $\varnothingcond$ denotes the null condition. Using the guidance-scaled output $\gtheta^w$, the self-consistency target $\scon$ is constructed by simulating two consecutive steps of size $d$:
\begin{align}
\scon = \gtheta^w(\xt, t, c, d)/2 + \gtheta^w(\bm{x}'_{t+d}, t+d, c, d)/2,
\label{eq:consistency_target}
\end{align}
where $\bm{x}'_{t+d}=\xt+\gtheta^w(\xt, t, c, d)d$. These components then form a unified loss function:
\begin{align}
\mathcal{L}^S(\theta) = \E_{\substack{\xzero \sim \N,\, (\xone, c) \sim D \\ (t,d) \sim p(t, d)}} \Big[ \underbrace{\lVert \stheta(\xt, t, c, d=0) - \sv \rVert^2}_{\text{Flow-Matching}} + \underbrace{\lVert \stheta(\xt, t, c, 2d) - \scon \rVert^2}_{\text{Self-Consistency}} \Big].
\label{eq:shortcut}
\end{align}
This approach allows shortcut models to achieve effective multi-step, few-step, and single-step generation using a single trained network.


%% file: tex/method.tex
\section{Improved Guidance Sampling for Shortcut Models} \label{sec:method-guidance}
Shortcut models \cite{shortcut} represent a significant step toward efficient one-step, few-step, and many-step generation. However, their integration of CFG introduces unique challenges compared to standard diffusion methods. This section revisits the shortcut model formulation to identify and address two critical limitations: (1) \textit{inflexibility due to fixed guidance scales}, (2) an \textit{overlooked flaw of accumulation} when applying CFG.

\subsection{Problem Statement}


\textbf{Inflexibility of Fixed Guidance.} The original shortcut model framework requires using a fixed guidance scale $w$ during training. This approach is restrictive for two main reasons. First, selecting an optimal $w$ is a difficult hyperparameter tuning problem that depends on the model's final behavior and the intended number of inference steps. Second, and more critically, it removes a key control for balancing the fidelity-diversity trade-off at inference time, substantially reducing the model's \textit{flexibility in production settings}.

\begin{figure}
    \centering
    \includegraphics[width=1.0\linewidth]{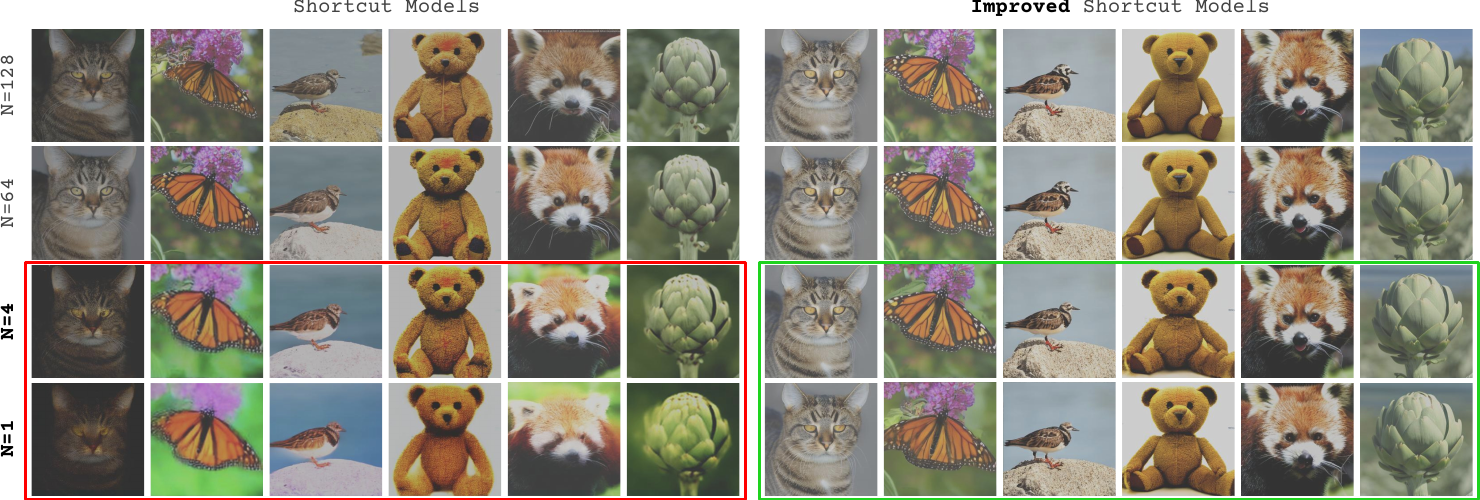}
    \vspace{0.00001em}
    \caption{Visualizing the \textit{accumulated guidance} problem in standard shortcut models (left) versus our integrated‐guidance approach (right). Top row shows multi-step generation; bottom row shows one-step generation. In the original shortcut model, the fixed guidance strength learned at training time effectively compounds across the many implicit sub-steps, leading to oversaturated colors and blurriness. By conditioning the network on the guidance scale and training it to apply guidance explicitly at each step, our method produces consistent, artifact-free images even in the extreme one-step and few-step regimes.}
    \vspace{-0.5cm}
    \label{fig:oversaturation}
\end{figure}

\textbf{Flaw of Accumulated Guidance.} A critical, previously overlooked issue in shortcut models is the accumulation of guidance effects. We are the first to identify and formalize this critical flaw, which stems from the recursive application of a fixed guidance scale. When a shortcut model generates a sample in a single large step, it implicitly combines the effects of many smaller, guided steps. This \textit{compounding characteristic} leads to an \textit{unintended amplification} of the guidance signal.

To formalize this, we consider a shortcut model $\stheta$ and its corresponding classifier-free guided output $\gtheta^w$ with guidance scale $w$. To analyze the generation process, we define a sequence of intermediate points $\{\bm{x}'_{\frac{i}{N}}\}_{i=0}^N$. This sequence is constructed by recursively applying $N$ consecutive shortcut steps, each with the smallest size $d=1/N$. The process begins from the initial noise $\xzero$ at $t=0$, conditioned on $c$. The starting point of the sequence is thus defined as $\bm{x}'_0 = \xzero$, and subsequent intermediate points are generated as follow:
\begin{equation}
    \bm{x}'_{\frac{i+1}{N}} = \bm{x}'_\frac{i}{N} + \stheta\left(\bm{x}'_\frac{i}{N}, \frac{i}{N}, c, d\right)d, \quad \text{for } i=0, \dots, N-1.
\end{equation}
\propositionbox{
\begin{proposition}\label{prop:accum_cfg}
    The model's prediction for a single large shortcut step of size $Nd=1$ approximately equals the average of the guided displacements corresponding to the $N$ smallest steps, but with an exponentially compounded guidance scale:
\begin{equation} \label{eq:accum_cfg_final}
    \stheta(\xzero, 0, c, Nd) \approx \frac{1}{N} \sum_{i=0}^{N-1} \gtheta^{w^{\log_2(N)}} \left(\bm{x}'_{\frac{i}{N}}, \frac{i}{N}, c, d \right).
\end{equation}
\end{proposition}
\par\noindent\textit{Proof.} See Appendix A.\hfill\qedsymbol
}

The original shortcut model \cite{shortcut} defines a fixed CFG scale $w=1.5$, exclusively applied when constructing self-consistency targets in \cref{eq:consistency_target}. Consequently, the one-step generation $\stheta(\xzero,0,c,1)$ implicitly aggregates the cumulative effect of $N$ intermediate steps. As formalized in Proposition \ref{prop:accum_cfg}, the guidance scale at each of these implicit intermediate steps is not $w$ but a much higher value $w' = w^{\log_2(N)}$. For $N=128$ base steps and $w=1.5$, this calculation yields an extremely high intermediate guidance scale for single large shortcut step cases ($w' = 1.5^{\log_2(128)}\approx 17$). Such high implicit guidance causes artifacts like \textit{over-saturation}, particularly in one-step and few-step generations (see \cref{fig:oversaturation}).
\subsection{Intrinsic Guidance Training for Shortcut Models}

To address these limitations of shortcut models \cite{shortcut}, we introduce an \textit{Intrinsic Guidance} training framework. In contrast to CFG implementations that operate between model outputs, as commonly practiced in multi-step diffusion models, we integrate the guidance mechanism into the \textit{network's internal space}. The core idea is to make the guidance scale $w$ an \textit{explicit input} to the model and to train the model to produce \textit{guided outputs directly} across a range of scales. This framework enables flexible CFG for few-step and many-step generation and, importantly, provides a principled way to apply CFG for one-step generation while resolving the accumulated guidance issue.

\textbf{Model Parameterization.}
We define the velocity field generator as a neural network $\stheta$. Unlike previous shortcut models \cite{shortcut}, $\stheta$ receives the CFG scale $w \ge 0$ as an additional input, alongside the current sample $\xt$, time $t$, condition $c$, and the target step size $d$. The network is thus trained to directly output the CFG-modulated velocity $\stheta(\xt, t, c, d, w)$, allowing guided sample generation with a single network evaluation per step. For implementation details, please refer to Appendix C.

\textbf{Flow Matching Objective.}
This objective \cite{lipman2022flow, liu2022flow} guides the model to predict the conditional and unconditional velocity for infinitesimal steps ($d=0$) without guidance ($w=0$), establishing the base vector fields upon which guidance will be built:
\begin{align}
    \begin{gathered}
        \Ls_{\velocity}(\theta) \coloneqq \E_{\substack{\xzero \sim \N,\, (\xone, c) \sim D \\ t \sim p(t)}} \Big[ \| \stheta(\xt, t, c, d=0, w=0) - \sv \|^2 \Big], \\
        \text{where} \quad \xt = (1-t)\xzero + t\xone \quad \text{and} \quad \sv = \xone - \xzero.
        \label{eq:velocity_loss} 
    \end{gathered}
\end{align}
When sampling the condition $c$ for training, we randomly include the null condition $\varnothingcond$ via stochastic dropout, following the standard scheme \cite{ma2024sit}.

\textbf{Intrinsic Guidance Objective.} This objective guides the model to directly produce the CFG-scaled output for infinitesimal steps ($d=0$) when being conditioned on non-zero guidance strengths ($w>0$). 
We begin by recalling the standard formulation of CFG with $w > 0$:
\begin{align}
    \stheta(\cdot, c, d=0, w) \approx \stheta(\cdot, c, d=0, w=0) + w \cdot \underbrace{(\stheta(\cdot, c, d=0, w=0) - \stheta(\cdot, \varnothingcond, d=0, w=0))}_{\sg},
    \label{eq:cfg_standard} 
\end{align}
where $\sg$ is the estimated guidance direction using the model's base prediction ($w=0$).

We substitute the unguided velocity $\stheta(\cdot,c,d=0,w=0)$ in \cref{eq:velocity_loss} based on the approximation of $\stheta(\cdot,c,d=0,w=0)$ in \cref{eq:cfg_standard}:
\begin{align*} 
     \stheta(\cdot, c, d=0, w=0)  &\approx  \stheta(\cdot, c, d=0, w) - w \cdot \sg, \\
    \| \stheta(\cdot, c, d=0, w=0) - \sv \|^2 &\approx \| (\stheta(\cdot, c, d=0, w) - w \cdot \sg) - \sv \|^2 \\
    &= \| \stheta(\cdot, c, d=0, w) - (\sv + w \cdot \sg) \|^2.
\end{align*}
Integrating this derived target into \cref{eq:velocity_loss}, we define a new training objective called the intrinsic guidance loss $\Ls_{\guidance}$. Crucially, we apply the stop-gradient operator  $\stopgrad{\cdot}$ to the guidance direction $\sg$ within this loss. This \textit{isolates the optimization} to focus solely on how the network learns to scale its output in response to $w > 0$, preventing interference with the optimization of the foundational $w=0$ predictions in $\mathcal{L}_{\velocity}$ and improving training stability.
\begin{align}
    \Ls_{\guidance}(\theta) \coloneqq \E_{\substack{\xzero \sim \N,\, (\xone, c) \sim D \\ (t, w) \sim p(t, w)}} \Big[ \| \stheta(\xt, t, c, d=0, w) - (\sv + w \cdot \stopgrad{\sg}) \|^2 \Big],
    \label{eq:guidance_loss} 
\end{align}
where guidance scale $w > 0$ is drawn from a distribution $p(w)$. This loss trains the network to learn how CFG works for infinitesimal steps ($d=0$).

\textbf{Guided Self-Consistency Objective.} This objective generalizes the self-consistency principle from \cite{shortcut} to operate with arbitrary step sizes ($d>0$) and any guidance scale ($w \ge 0$). The objective maintains the foundational properties of shortcut models, where a \textit{single, large guided shortcut step} yields an output consistent with the composition of \textit{two smaller, consecutive guided steps}.
\begin{align}
    \begin{gathered}
        \Ls_{\consistency}(\theta) \coloneqq \E_{\substack{\xzero \sim \N,\, (\xone, c) \sim D \\ (t, w, d) \sim p(t, w, d)}} \Big[ \| \stheta(x_t, t, c, 2d, w) - s_\mathrm{consistency} \|^2 \Big], \\
        \quad\text{where}\quad
        s_\consistency \coloneqq s_{\theta^{-}}(x_t, t, c, d, w)/2 + s_{\theta^{-}}(x'_{t+d}, t, c, d, w)/2 \\ 
        \text{and} \quad x'_{t+d} = x_t + s_\theta(x_t, t, c, d, w)d, \quad 
        \label{eq:consistency_loss}
    \end{gathered}
\end{align}
where $\theta^{-}$ is the EMA target network. The stop-gradient operator $\stopgrad{\cdot}$ is applied to the entire consistency target to stabilize training, following standard practice for self-consistency objectives.

\textbf{Final Training Objective.} The final objective is a weighted sum of the individual components as follows:
\begin{align}
    \Ls_{\total}(\theta) = \alpha \Ls_{\velocity}(\theta) + \beta \Ls_{\guidance}(\theta) + \gamma \Ls_{\consistency}(\theta) ,
    \label{eq:total_loss}
\end{align}
where $\alpha, \beta, \gamma > 0$ are hyperparameters. For simplicity, we set $\alpha = \beta = \gamma = 1$ in our experiments.

\subsection{Interval Guidance in Training} 

Findings presented in \cite{interval} suggest that during the inference of diffusion models, strong classifier-free guidance can be \textit{detrimental during the early stages} of the reverse process, corresponding to high noise levels (time $t$ approaching 0 under the convention where $t=0$ is pure noise).

At high noise levels, the signal-to-noise ratio is low. Hence, the unconditional prediction $\stheta(\xt, t, \varnothingcond, d = 0, w=0)$ points towards the average image in the dataset $D$, while the conditional prediction $\stheta(\xt, t, c, d=0, w=0)$ points vaguely towards the average image for that condition. Consequently, the guidance vector $\sg$ points roughly from the global data mean towards the conditional data mean. Applying strong extrapolation ($w>0$) along this coarse direction to a high-entropy state $\xt$ can cause the sample diversity to \textit{collapse prematurely} towards a few dominant modes.

To mitigate this issue, we incorporate this insight into the training objective $\Ls_{\guidance}$ in \cref{eq:guidance_loss} by applying guidance only within a defined time interval $[ t_{\interval}, 1)$. In particular, we modify the guidance scale $w$ used within the \cref{eq:guidance_loss} to be time-dependent:
\begin{align}
    w(t) &~=~
      \begin{cases}
        w, & \text{if } t \in \left[t_{\interval},1 \right) \\
        0, & \text{otherwise,}
      \end{cases}
      \label{eq:cfg_interval_training}
\end{align}
where $w$ is the sampled guidance scale. This strategy trains the model to apply zero guidance at high noise levels, preserving diversity in the critical early phase of generation. We find that $t_{\interval}=0.3$ works well empirically, and this threshold is represented by the red dashed line in \cref{fig:self_consistency_loss,fig:flow_matching_loss}. Notably, in the high-noise regime where $t<0.3$ (i.e., before guidance is applied), both flow-matching and self-consistency losses exhibit lower loss pattern.


\section{Multi-Level Wavelet Function Against Frequency Bias} \label{sec:method-wavelet}
Deep neural networks often prioritize learning \textit{low-frequency information over high-frequency details}, a phenomenon known as frequency bias \cite{spectralbias,freqbias,basri2020frequency,xu2019frequency,xu2019training}. This bias is detrimental to generative models \cite{chen2021ssd,frank2020leveraging,khayatkhoei2022spatial,schwarz2021frequency}, particularly shortcut models \cite{shortcut} optimized for few-step generation with low-level direct domain losses, such as the $\ell_2$ loss. These models typically capture global structure but lack fine-grained, high-frequency details, resulting in blurred textures. The low-level $\ell_2$ loss contributes to this, as it is often dominated by errors in low-frequency components \cite{zhang2023hipa}.

To mitigate this frequency bias, we introduce optimizing the reconstruction objective in the \textit{wavelet domain}. Specifically, discrete wavelet transform ($\DWT$) is employed to decompose both the prediction and the target (\cref{eq:velocity_loss,eq:guidance_loss,eq:consistency_loss}) into their respective wavelet coefficients, yielding multi-band representations, which introduce a \textit{frequency-aware error signal}. By isolating reconstruction errors at different frequency levels, the optimization process is better guided to preserve high-frequency details that are often neglected in standard losses. Consequently, the model learns to generate outputs with significantly improved perceptual quality, capturing both the global structure and fine-grained details.

To further refine supervision across the frequency spectrum, we extend this to a multi-level $\DWT$ decomposition. This process involves recursively decomposing the wavelet sub-bands from the previous level. Our experiments consistently show that using a deeper multi-level decomposition, with five levels ($L=5$) in our experiments, substantially improves the synthesis of high-frequency content compared to a single-level $\DWT$ loss or standard pixel-wise objectives. Pseudocode is provided in Appendix B.



\section{Scaling Optimal Transport for Better Trajectories}
\label{sec:method-sOT}

\begin{figure}[t]
    \vspace{-1.0cm}
    \begin{subfigure}[t]{0.39\textwidth}
        \centering 
        \begin{adjustbox}{width=\linewidth, valign=T} 
            \includegraphics{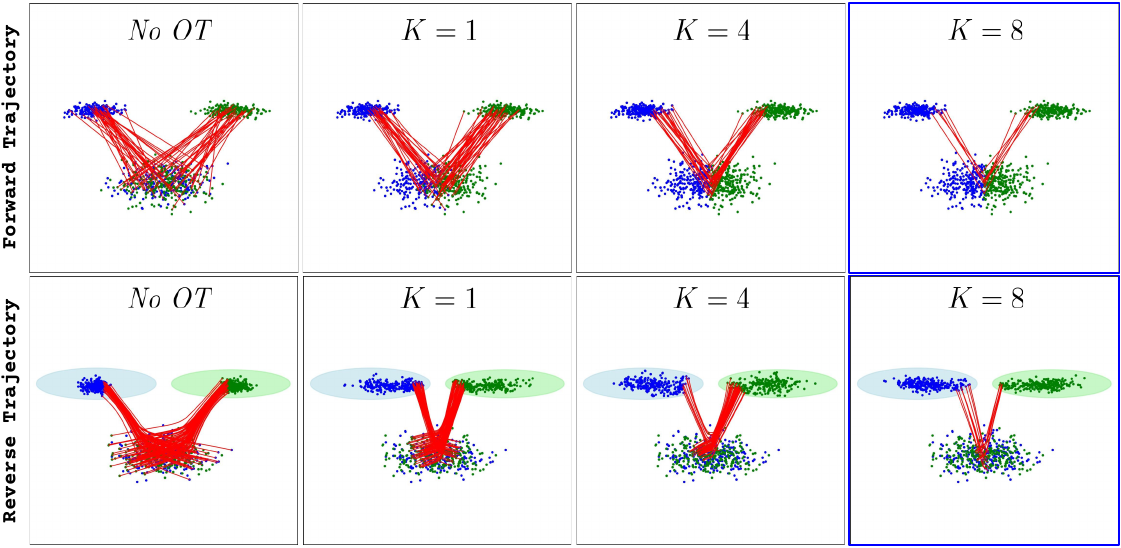} 
        \end{adjustbox}
        \vspace{0.35cm}
        \vspace{-0.2cm}
        \caption{Forward and reverse trajectories.}
        \label{fig:sot_trajectories}
    \end{subfigure}
    \hfill 
    \begin{subfigure}[t]{0.29\textwidth}
        \centering 
        \begin{adjustbox}{width=\linewidth, valign=T} 
            \includegraphics{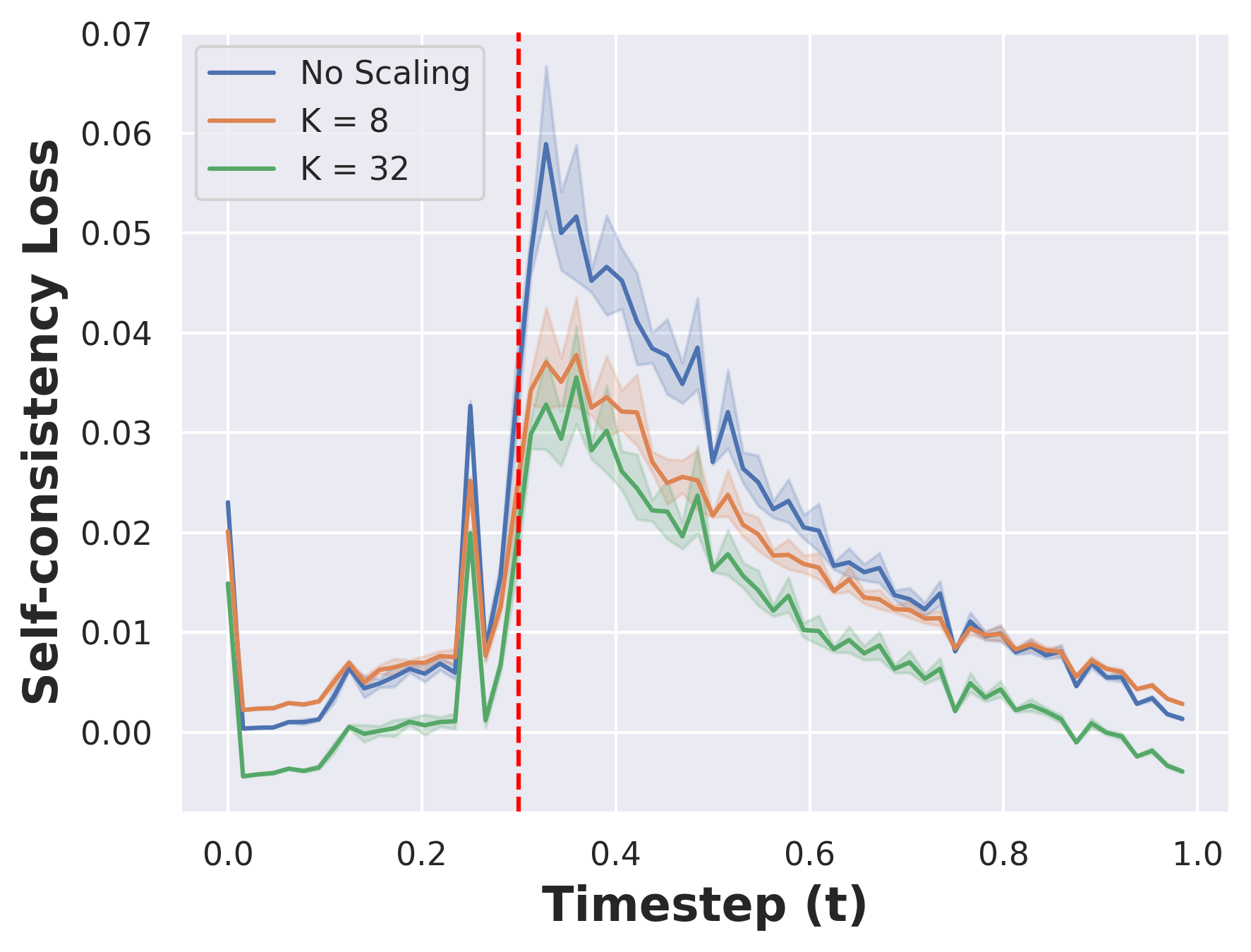} 
        \end{adjustbox}
        \vspace{0.02cm}
        \vspace{-0.2cm}
        \caption{\centering Self-consistency Loss.} 
        \label{fig:self_consistency_loss}
    \end{subfigure}
    \hfill 
    \begin{subfigure}[t]{0.29\textwidth}
        \centering 
        \begin{adjustbox}{width=\linewidth, valign=T} 
            \includegraphics{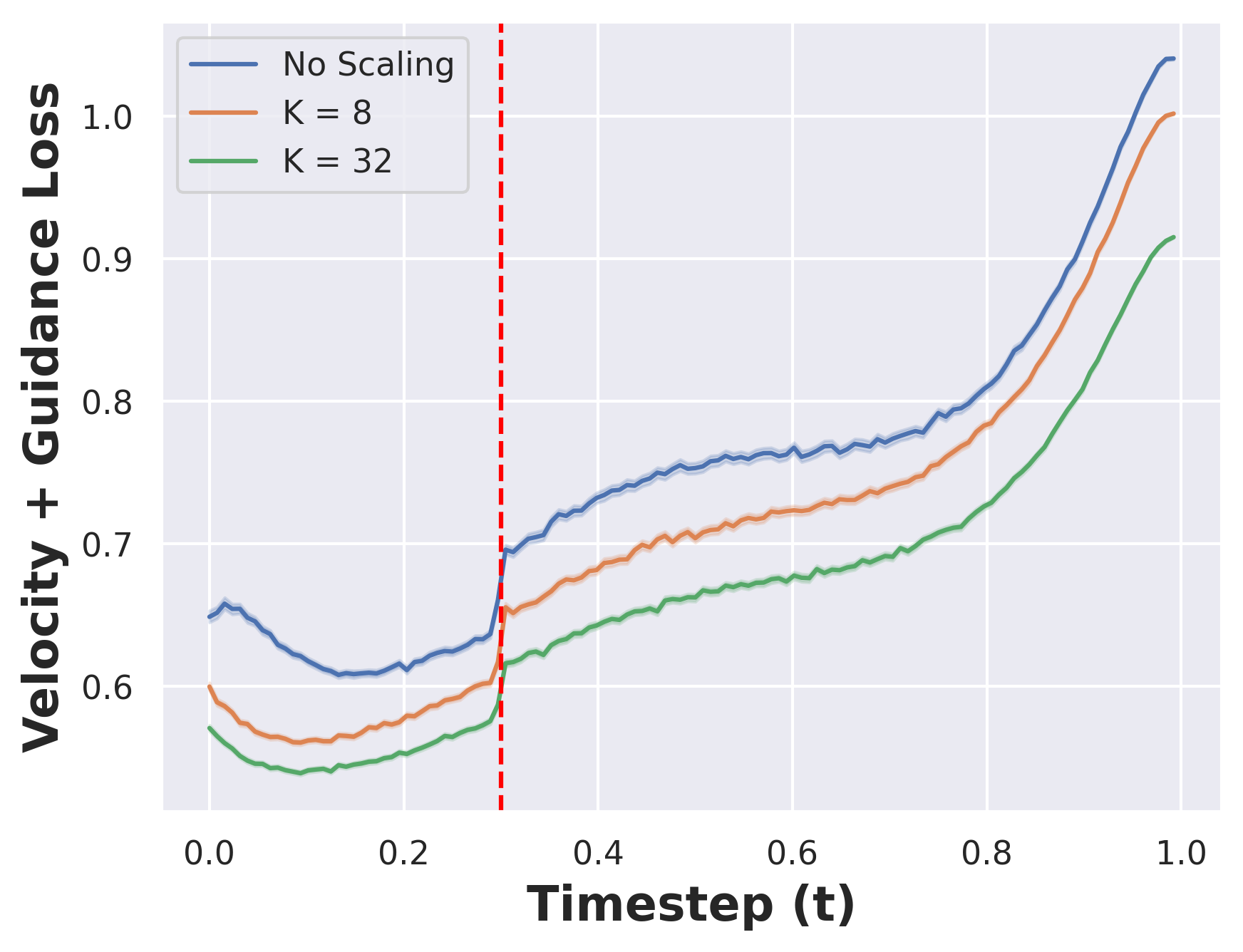} 
        \end{adjustbox}
        \vspace{0.02cm}
        \vspace{-0.2cm}
        \caption{\centering Velocity \& Guidance Losses.}
        \label{fig:flow_matching_loss}
    \end{subfigure}
    \vspace{0.4cm}
    \caption{\textbf{Efficacy of Scaling Optimal Transport (sOT) Matching} in improving shortcut model training, demonstrated by varying the OT scaling factor $K$. \textbf{(a)} On a bimodal target, forward trajectories (top row) without OT exhibit \textit{frequent intersections} (red), compelling the reverse generative process (bottom row) to follow \textit{high-curvature paths} that initially average the target modes (blue, green).
    Our sOT approach, by increasing the effective OT scaling $K$, \textit{progressively disentangles these forward couplings}, yielding \textit{substantially straighter reverse trajectories}. \textbf{(b, c)} The benefits of this trajectory straightening are reflected in the training losses: both self-consistency (b) and velocity \& guidance (c) losses are \textit{consistently lower and more stable with larger $K$}, especially after the interval guidance point $t_{\interval}=0.3$ (left-side of the red dashed line). \textbf{Note for (a)}: To avoid overly dense trajectory, only intersecting forward paths and sampling trajectories with high curvature are highlighted. Zoom for detail.} 
    \vspace{-0.4cm}
    \label{tab:ot_figure} 
    \centering 
\end{figure}


Shortcut models, similar to standard flow matching, often encounter considerable loss variance during training. This instability, as highlighted in prior works \cite{pooladian2023multisample, lee2023minimizing}, primarily stems from the \textit{conventional random pairing} of initial noise samples $\xzero \sim \mathcal{N}(0, \mathbf{I})$ with target images ($\xone \sim D$). Such random coupling frequently results in \textit{intersecting forward trajectories} ($\xt$), where similar noise inputs might be mapped to vastly different data targets. Consequently, the learned generative ODE exhibits \textit{high-curvature reverse paths}, posing a significant challenge for shortcut models. Optimal Transport (OT) within mini-batches can mitigate this by re-assigning noise-image pairs to minimize a transport cost, typically $L_2$ distance, yielding smoother trajectories \cite{pooladian2023multisample, lee2023minimizing}. 



Although mini-batch OT helps, we recognize that the efficacy of such mini-batch OT is fundamentally linked to the number of samples considered. Theoretical and empirical findings \cite{cOT} show that OT better approximates the true transport plan as the number of samples $n$ increases. However, scaling the mini-batch size $M$ to achieve this is often limited by GPU memory constraints during training. To tackle this scalability challenge, we introduce Scaling Optimal Transport (sOT), a strategy that \textit{decouples the effective batch size for OT computation from the batch size used for model training}. Our method allows us to use a large number of samples for OT without increasing the memory needed for model training. 
Specifically, for every $K$ training batches of size $M$, we aggregate all noise and image samples into a single set of size $K \times M$ and compute one OT plan over it. The resulting $K \times M$ matched pairs are then split back into $K$ mini-batches of size $M$ and used for the next $K$ standard training steps. This strategy allows us to benefit from an effective OT batch size of $K \times M$ while maintaining a computationally tractable batch size $M$ for model training. The computational overhead from sOT is modest, adding approximately $4\%$ to the total training time in our experiments.

\cref{fig:sot_trajectories} demonstrates sOT's effectiveness on a 2D bimodal dataset. Random pairing (leftmost column) creates overlapping forward trajectories, forcing the reverse process to follow high-curvature paths that initially move toward an average direction before bending toward target modes. As the effective OT batch size $K$ increases (columns 2-4), sOT produces progressively more \textit{disentangled forward couplings} with fewer intersections, resulting in \textit{straighter generative trajectories}.

\cref{fig:self_consistency_loss,fig:flow_matching_loss} further illustrate the benefits of sOT by showing a consistent decrease in self-consistency and flow-matching losses, as the sOT scaling factor $K$ increases, with these loss statistics computed over 1.2M ImageNet samples. Regarding self-consistency loss (\cref{fig:self_consistency_loss}), sOT \textit{straightens the learned generative paths} (as seen in \cref{fig:sot_trajectories}). On these smoother trajectories, a single large shortcut step \textit{more accurately approximates the outcome of two composed smaller steps}, thus reducing the discrepancy and lowering the loss. This indicates that the model learns more reliable long-range predictions. For flow-matching and guidance loss (\cref{fig:flow_matching_loss}), sOT provides more structured noise-data pairings. This results in a less conflicting target velocity field for the model to learn, as similar noise inputs are mapped to more consistently related data targets. This simplified regression task allows the model to learn a more coherent target, thereby reducing the error.

\section{Twin EMA Improves Self-Consistency Property} \label{sec:method-twinEMA}
Self-consistency objectives seek to align large-step generations with sequences of smaller steps. Specifically, they train an online network $\theta$ to produce outputs from a $2d$-sized step that match the consistency target $\scon$, which is obtained by applying two consecutive $d$-sized steps. Instead of using the online network $\theta$ to compute $\scon$, the original paper suggests using a target network $\theta^{-}$, typically implemented as an EMA of $\theta$. When utilized for generating self-consistency targets, EMA parameters provide crucial stability that \textit{dampens the oscillatory behaviors} that would otherwise propagate and amplify across different timesteps.

Despite its essential stabilizing role, conventional EMA implementations introduce a \textit{tension in self-consistency training}. When the target network $\theta^{-}$ is maintained with a standard slow decay rate, it inevitably represents a historical state of the online network rather than its target distribution. This \textit{temporal lag} means that self-consistency targets $\scon$ are derived from \textit{outdated networks}.

\input{table/main_table}
Consequently, the online network $\theta$ faces a \textit{conflicting objective}: it must simultaneously optimize for its current trajectory during many-step generation (i.e., flow matching loss) while aligning with targets generated from its historical states during few-step (i.e, self-consistency loss). As training progresses, this misalignment forces the model to learn and balance multiple, potentially contradictory generative mappings rather than converging toward a unified process. The slower the EMA decay rate, the more pronounced this distributional divergence becomes.

To resolve this tension while preserving the essential stabilizing benefits of EMA, we introduce a Twin EMA approach that maintains \textit{two distinct parameter sets}: (1) \textbf{Inference Parameters} ($\theta^{-}_{\infer}$): Updated with a conventional slow decay rate and used exclusively during inference, ensuring high-quality sample generation benefits from long-term parameter averaging. (2) \textbf{Target Parameters} ($\theta^{-}_{\target}$): Updated with a significantly faster decay rate to maintain close proximity to the online network's current state while still providing stabilization.
This \textit{decoupling addresses the fundamental conflict} between stability and recency requirements. By generating self-consistency targets from a near-contemporary version of itself, the online network more effectively enforces consistency across different timestep discretizations. Meanwhile, the separate maintenance of slow-decay parameters for inference preserves the stability crucial for high-quality sample generation.

%% file: table/main_table.tex
\begin{wraptable}{r}{6.7cm}
        \centering
        \captionsetup{font=small}
        \scriptsize
        \caption{\textbf{Quantitative results} on class-conditional ImageNet $256\times 256$ across different model types.}
        \vspace{4pt}
    
        \begin{tabular}{l c c c}
                \toprule
                Model & FID-50k ($\downarrow$) & NFE ($\downarrow$) & \#Params\\
                
                \midrule 
                \multicolumn{4}{c}{\textbf{GAN}}\\
                \midrule
                
                BigGAN-deep \citep{brock2018large} & 4.06 & 1 & 112M \\
                GigaGAN~\citep{kang2023scaling} & 3.45 & 1 & 569M \\
                StyleGAN-XL~\citep{karras2020analyzing} & 2.30 & 1 & 166M \\
                
                \midrule 
                \multicolumn{4}{c}{\textbf{Masked \& AR}}\\
                \midrule
                
                VQGAN~\citep{esser2021taming} & 26.52 & 1024 & 227M \\
                MaskGIT~\citep{chang2022maskgit} & 6.18 & 8 & 227M \\
                VAR-$d20$~\citep{tian2024visual} & 2.57 & 10 & 600M \\
                VAR-$d30$~\citep{tian2024visual} & 1.92 & 10 & 2B \\
                MAR~\citep{li2024autoregressive} & 1.98 & 100 & 400M \\
                
                \midrule
                \multicolumn{4}{c}{\textbf{Diffusion \& Flow Models}}\\
                \midrule
                
                ADM~\citep{dhariwal2021diffusion} & 10.94 & 250 & 554M \\
                CDM~\citep{ho2022cascaded} & 4.88 & 8100 & - \\
                SimDiff~\citep{hoogeboom2023simple} & 2.77 & 512 & 2B \\
                LDM-4-G~\citep{rombach2022high} & 3.60 & 250 & 400M \\
                U-DiT-L~\citep{tian2024u} & 3.37 & 250 & 916M \\
                U-ViT-H~\citep{bao2023all} & 2.29 & 50 & 501M \\
                DiT-XL/2 ~\citep{peebles2023scalable} & 2.27 & 250 & 675M \\
                SiT-XL/2 ~\citep{ma2024sit} & 2.15 & 250 & 675M \\
                REPA-XL/2 ~\citep{yu2025repa} & 1.42 & 250 & 675M \\
                FlowDCN ~\citep{flowdcn} & 2.00 & 250 & 618M \\
                \midrule 
                \multicolumn{4}{c}{\textbf{One-to-Many Step Models}}\\
                \midrule

                iCT~\citep{song2023consistency} & 34.24 & 1 & 675M \\
                & 20.3 & 2 & 675M \\
                SM~\citep{shortcut} & 10.60 & 1 & 675M \\
                & 7.80 & 4 & 675M \\
                & 3.80 & 128 & 675M \\
                IMM ~\citep{imm} & 7.77 & 1 & 675M \\
                & 3.99 & 2 & 675M \\
                & 2.51 & 4 & 675M \\
                & 1.99 & 8 & 675M \\
                \textbf{iSM (ours)} & 5.27 & 1 & 675M \\
                 & 2.44 & 2 & 675M \\
                 & 2.05 & 4 & 675M \\
                 & 1.93 & 8 & 675M \\
                & 1.88 & 128 & 675M \\
                
                \bottomrule
        \end{tabular}%
            \label{tab:main_table}
        \vspace{-1.0cm}
    \end{wraptable}

%% file: tex/experiment.tex
\section{Experiment} \label{sec:experiment}
\begin{wraptable}{r}{5.5cm}
    \centering
    \vspace{-13pt}
    \caption{\textbf{Broader evaluation} with FD-DINOv2 and IS on ImageNet $256\times256$.}
    \label{tab:additional_metrics}
    \scriptsize
    \vspace{6pt}
    \begin{tabular}{llcc}
        \toprule
        Model & NFE & FD-DINOv2$\downarrow$ & IS$\uparrow$ \\
        \midrule
        SM \cite{shortcut} & 1 & 500.92 & 102.66 \\
        IMM \cite{imm} & 1 & 247.78 & 128.87 \\
        \cellcolor{gray!15}\textbf{iSM (ours)} & \cellcolor{gray!15}1 & \cellcolor{gray!15}\textbf{232.31} & \cellcolor{gray!15}\textbf{223.52} \\
        \midrule
        SM \cite{shortcut} & 2 & 329.53 & 125.66 \\
        IMM \cite{imm} & 2 & 152.08 & 173.66 \\
        \cellcolor{gray!15}\textbf{iSM (ours)} & \cellcolor{gray!15}2 & \cellcolor{gray!15}\textbf{107.63} & \cellcolor{gray!15}\textbf{302.29} \\
        \midrule
        SM \cite{shortcut} & 4 & 265.90 & 136.79 \\
        IMM \cite{imm} & 4 & 110.88 & 204.95 \\
        \cellcolor{gray!15}\textbf{iSM (ours)} & \cellcolor{gray!15}4 & \cellcolor{gray!15}\textbf{83.70} & \cellcolor{gray!15}\textbf{298.23} \\
        \bottomrule
    \end{tabular}
    \vspace{-0.4cm}
\end{wraptable}

We combine all the improvements detailed from \Cref{sec:method-guidance,sec:method-wavelet,sec:method-sOT,sec:method-twinEMA} to train our improved shortcut models. We extensively benchmark our models on ImageNet $256 \times 256$ against various models across multiple training paradigms. We evaluate sample quality using FID-50K \cite{fid}. For all experiments, our models use the XL/2 variant of the SiT architecture \cite{ma2024sit}. We report evaluations against other models at \textit{800K iterations}, while component-wise analyses are conducted at \textit{250K iterations}. We summarize our main quantitative benchmarks in \Cref{tab:main_table}, provide a broader evaluation using additional metrics in \Cref{tab:additional_metrics}, present a detailed component-wise analysis in \Cref{tab:ablation}, and demonstrate the scalability and generality of our framework in \Cref{tab:scalability}.
\subsection{Quantitative result and comparison}\label{sec:results}

Our model, iSM, achieves strong FID scores across various NFE settings, demonstrating high sample quality and inference efficiency. With 8 inference steps, iSM \textit{surpasses the 10-step VAR} \citep{tian2024visual} of similar model size and performs \textit{comparably to VAR's much larger 2B variant}. A key advantage of iSM is its inherent support for \textit{flexible NFE counts}—from single-step to many-step generation—a feature absent in models like VAR, with FID scores significantly improving as NFE increases from 1 to 8. Furthermore, iSM performs favorably against other recent variable-step models, including the original shortcut model \citep{shortcut} and IMM \citep{imm}. For instance, in single-step generation, iSM attains an FID of 5.27, \textit{outperforming both}. This strong performance extends to few-step scenarios, with FIDs of 2.05 (4 steps) and 1.93 (8 steps), highlighting iSM's robustness across different inference budgets.

To provide a more comprehensive validation, we extend our evaluation to include FD-DINOv2 and Inception Score (IS). As presented in Table~\ref{tab:additional_metrics}, our iSM framework consistently outperforms the baselines across these metrics. The strong FD-DINOv2 scores confirm that our model's high fidelity is robust and not specific to the Inception feature space; for example, at 4 NFE, iSM achieves a score of 83.70, a \textit{greater than 3 times improvement} over the baseline SM's 265.90. The Inception Scores show a particularly large margin of improvement; for example, at 2 inference steps, our model's IS of 302.29 more than \textit{doubles} the baseline score of 125.66. The consistent gains across these distinct evaluation metrics provide strong evidence that iSM yields significant and robust improvements to shortcut model generation.

\subsection{Component-wise Analysis}\label{sec:ablation}

We explore whether our proposed techniques enhance shortcut model training. As shown in \cref{tab:ablation}, each component contributes to improved generation quality, with their combination achieving a significantly better FID score compared to the vanilla model. Below, we present a detailed analysis of the individual impact of each component.

\input{table/ablation}
\sethlcolor{cyan!15}
\hl{\textbf{Intrinsic Guidance.}}
We examine the effect of varying the range of the conditioning CFG scale $w$ on model performance. During training, we sample discrete values of $w$ from the interval $[0, w_{\wmax}]$ using a step size of 0.25. Through empirical evaluation, we find that setting $w_{\wmax} = 3.5$ leads to the best results. When $w_{\wmax}$ is set too high (e.g., 5.0), the model is exposed to a broader range of conditioning strengths. This can introduce unnecessary complexity, as the model must learn to generalize over a wider set of conditioning values, potentially degrading overall performance. On the other hand, if $w_{\wmax}$ is too low (e.g., 2.0), the model only learns from outputs corresponding to low CFG scales, which may lack high quality.

\sethlcolor{red!15}
\hl{\textbf{Interval Guidance in Training.}} We examined the effect of varying $t_{\interval}$, the threshold below which CFG is disabled. We observed that increasing $t_{\interval}$ from very low values up to a moderate range (e.g., $0.1$ to $0.3$) improved both metrics, consistent with the benefit of avoiding strong, coarse guidance in high-noise regimes. However, further increasing $t_{\interval}$ to higher values (e.g., $0.5$) results in a significant increase in FID. Training the model without guidance for a substantial portion of the early reverse process may impair its ability to effectively leverage guidance when $t \ge t_{\interval}$.

\sethlcolor{yellow!15}
\hl{\textbf{Multi-level Wavelet Function.}} We also investigate the impact of varying the number of decomposition levels in the wavelet transform. Given that the latent representation is of size $32 \times 32$, the maximum feasible number of levels is $\log_2(32) = 5$. Our experiments show that utilizing the full 5 levels yields better performance compared to using fewer levels (e.g., 1 or 3) or not using the wavelet transform entirely.

\sethlcolor{green!15}
\hl{\textbf{Scaling Optimal Transport.}}
We also examine the impact of OT matching on the performance of our previous improvements. Our results show that using traditional OT matching results in lower FID scores than not using OT. We then assess the effectiveness of our proposed enhanced OT by applying it at two different scales: $K=8$ and $K=32$. Our results reveal a clear correlation between increasing OT scale and improved FID scores. Although scaling beyond $K=32$ may further enhance performance, we select $K=32$ as a trade-off between computational overhead and effectiveness.

\sethlcolor{blue!15}
\hl{\textbf{Twin EMA.}} Finally, our proposed Twin EMA was evaluated using decay rates of $0.999$ and $0.95$ for the target network $\theta^{-}_{\text{target}}$. A decay rate of $0.95$ provided an effective balance between training stability and mitigating distributional divergence from self-consistency targets. This was evidenced by a substantial reduction in FID, achieving 6.56 and 2.16 for one-step and four-step generation, respectively.

\subsection{Generalizing to Higher Resolutions and Architectures}
\label{sec:scalability}

\begin{wraptable}{r}{6.0cm}
    \centering
    \vspace{-13pt}
    \caption{\textbf{Scalability and generality} of iSM on ImageNet $512\times512$ with the FlowDCN architecture, reported after 300K training iterations.}
    \label{tab:scalability}
    \vspace{4pt}
    \scriptsize
    \begin{tabular}{llccc}
        \toprule
        Model & NFE & FID$\downarrow$ & Precision$\uparrow$ & Recall$\uparrow$ \\
        \midrule
        SM & 1 & 43.81 & 0.56 & 0.11 \\
        \cellcolor{gray!15}\textbf{iSM (ours)} & \cellcolor{gray!15}1 & \cellcolor{gray!15}\textbf{37.05} & \cellcolor{gray!15}\textbf{0.60} & \cellcolor{gray!15}\textbf{0.55} \\
        \midrule
        SM & 4 & 12.16 & \textbf{0.86} & 0.19 \\
        \cellcolor{gray!15}\textbf{iSM (ours)} & \cellcolor{gray!15}4 & \cellcolor{gray!15}\textbf{9.94} & \cellcolor{gray!15}0.78 & \cellcolor{gray!15}\textbf{0.62} \\
        \bottomrule
    \end{tabular}
    \vspace{-0.4cm}
\end{wraptable}

To demonstrate the \textit{architectural generality} and \textit{resolution scalability} of our framework, we applied our \textbf{iSM} training to FlowDCN~\cite{wang2024exploring}, a fully convolutional generative model with group-wise deformable convolution blocks, on ImageNet $512\times512$. As shown in Table~\ref{tab:scalability}, at 300K training iterations, our method yields a substantial FID improvement over the baseline, confirming its effectiveness across different model families and at higher scales. For instance, in 4-step generation, our approach reduces the FID from 12.16 to 9.94. The improvement is also pronounced in the one-step generation, where iSM improves sample diversity by \textit{boosting Recall over 5 times} from 0.11 to 0.55.

%% file: table/ablation.tex
\begin{table*}[t]
\centering
\vspace{-3pt}
\caption{\textbf{Ablation study} on ImageNet $256\times256$. We investigate the impact of key hyperparameters for each of our proposed components. The best-performing setting from each block is carried forward to the next. All models are trained for 250K iterations.}
\vspace{4pt}
\label{tab:ablation}
\scriptsize
\begin{tabular}{l c c c c c c c}
    \toprule
    Method & $w_{\wmax}$ & $t_{\interval}$ & $L$ & $K$ & $\theta^{-}_{\text{target}}$ & FID$_{\one}$ $\downarrow$ & FID$_{\four}$ $\downarrow$ \\
    \midrule
    \multirow{3}{*}{Intrinsic Guidance (Sec. \ref{sec:method-guidance})} & \cellcolor{cyan!15} 2.0 & 0 &  0 & 0 & 0.9999 & 10.10 & 3.21 \\
    & \cellcolor{cyan!15} \textbf{3.5} & 0 &  0   & 0 & 0.9999 & 9.62  & 3.17 \\
    & \cellcolor{cyan!15} 5.0 & 0 &  0   & 0 & 0.9999 & 10.38 & 3.34 \\
    \midrule
    \multirow{4}{*}{Interval Guidance (Sec. \ref{sec:method-guidance})} & 3.5 & \cellcolor{red!15}0.0 &  0   & 0 & 0.9999 & 9.62 & 3.17 \\
    & 3.5 & \cellcolor{red!15}0.1 &  0   & 0 & 0.9999 & 8.58 & 3.14 \\
    & 3.5 & \cellcolor{red!15}\textbf{0.3} &  0   & 0 & 0.9999 & 8.49 & 2.81 \\
    & 3.5 & \cellcolor{red!15}0.5 &  0   & 0 & 0.9999 & 19.22 & 2.84 \\
    \midrule
    \multirow{4}{*}{Multi-level Wavelet (Sec. \ref{sec:method-wavelet})} & 3.5 & 0.3 & \cellcolor{yellow!15}0   & 0 & 0.9999 & 8.49 & 2.81 \\
    & 3.5 & 0.3 & \cellcolor{yellow!15}1   & 0 & 0.9999 & 8.21  & 2.79 \\
& 3.5 & 0.3 & \cellcolor{yellow!15}3   & 0 & 0.9999 & 8.17  & 2.75 \\
& 3.5 & 0.3 & \cellcolor{yellow!15}\textbf{5}   & 0 & 0.9999  & 8.12 & 2.64 \\
\midrule
\multirow{4}{*}{Scaling OT (Sec. \ref{sec:method-sOT})} & 3.5 & 0.3 & 5  & \cellcolor{green!15}0  & 0.9999 & 8.12 &  2.64 \\
& 3.5 & 0.3 & 5  & \cellcolor{green!15}1  & 0.9999  & 8.07 & 2.51  \\
& 3.5 & 0.3 & 5  & \cellcolor{green!15}8  & 0.9999 & 8.03  & 2.28  \\
& 3.5 & 0.3 & 5  & \cellcolor{green!15}\textbf{32} & 0.9999  & 7.97  & 2.23\\
\midrule
\multirow{3}{*}{Twin EMA (Sec. \ref{sec:method-twinEMA})} & 3.5 & 0.3 & 5  & 32 & \cellcolor{blue!15}0.9999 & 7.97 &  2.23 \\
& 3.5 & 0.3 & 5  & 32 & \cellcolor{blue!15}0.999 & 7.43 & 2.21\\
& 3.5 & 0.3 & 5  & 32 & \cellcolor{blue!15}\textbf{0.95} & 6.56 & 2.16  \\
\bottomrule
\vspace{-6pt}
\end{tabular}
\end{table*}

%% file: tex/conclusion.tex
\section{Conclusion} \label{sec:conclusion}

This work confronts the foundational obstacles that have previously limited the performance of shortcut models. We systematically address five core challenges: the damaging effects of compounding guidance, inflexible fixed guidance, a pervasive low-frequency bias, conflicts between self-consistency and EMA updates, and instability from high-curvature flow trajectories. To resolve these issues, we present \textbf{iSM}, a unified training framework composed of four key components. By incorporating \textit{Intrinsic Guidance} for dynamic control while resolving the flaw of compounding guidance, a \textit{Multi-Level Wavelet Loss} for high-frequency fidelity, \textit{Scaling Optimal Transport (sOT)} for smoother trajectories, and a \textit{Twin EMA} strategy for stable consistency, iSM elevates shortcut models into a competitive paradigm for generative modeling.

Our extensive experiments on ImageNet $256 \times 256$ demonstrate that iSM yields substantial FID improvements over baseline shortcut models across one-step, few-step, and multi-step generation. These results close the performance gap with leading GANs and other diffusion-based models, establishing shortcut models as a viable and competitive alternative that uniquely offers variable sampling budgets from a single network.

%% file: checklist.tex
\section*{NeurIPS Paper Checklist}

\begin{enumerate}

\item {\bf Claims}
    \item[] Question: Do the main claims made in the abstract and introduction accurately reflect the paper's contributions and scope?
    \item[] Answer: \answerYes{} 
    \item[] Justification: Our contributions are accurately reflected in the abstract and in the summary of introduction section \cref{sec:intro}.
    \item[] Guidelines:
    \begin{itemize}
        \item The answer NA means that the abstract and introduction do not include the claims made in the paper.
        \item The abstract and/or introduction should clearly state the claims made, including the contributions made in the paper and important assumptions and limitations. A No or NA answer to this question will not be perceived well by the reviewers. 
        \item The claims made should match theoretical and experimental results, and reflect how much the results can be expected to generalize to other settings. 
        \item It is fine to include aspirational goals as motivation as long as it is clear that these goals are not attained by the paper. 
    \end{itemize}

\item {\bf Limitations}
    \item[] Question: Does the paper discuss the limitations of the work performed by the authors?
    \item[] Answer: \answerYes{} 
    \item[] Justification: We explicitly mention the limitation for future works in \cref{sec:conclusion}.
    \item[] Guidelines:
    \begin{itemize}
        \item The answer NA means that the paper has no limitation while the answer No means that the paper has limitations, but those are not discussed in the paper. 
        \item The authors are encouraged to create a separate "Limitations" section in their paper.
        \item The paper should point out any strong assumptions and how robust the results are to violations of these assumptions (e.g., independence assumptions, noiseless settings, model well-specification, asymptotic approximations only holding locally). The authors should reflect on how these assumptions might be violated in practice and what the implications would be.
        \item The authors should reflect on the scope of the claims made, e.g., if the approach was only tested on a few datasets or with a few runs. In general, empirical results often depend on implicit assumptions, which should be articulated.
        \item The authors should reflect on the factors that influence the performance of the approach. For example, a facial recognition algorithm may perform poorly when image resolution is low or images are taken in low lighting. Or a speech-to-text system might not be used reliably to provide closed captions for online lectures because it fails to handle technical jargon.
        \item The authors should discuss the computational efficiency of the proposed algorithms and how they scale with dataset size.
        \item If applicable, the authors should discuss possible limitations of their approach to address problems of privacy and fairness.
        \item While the authors might fear that complete honesty about limitations might be used by reviewers as grounds for rejection, a worse outcome might be that reviewers discover limitations that aren't acknowledged in the paper. The authors should use their best judgment and recognize that individual actions in favor of transparency play an important role in developing norms that preserve the integrity of the community. Reviewers will be specifically instructed to not penalize honesty concerning limitations.
    \end{itemize}

\item {\bf Theory assumptions and proofs}
    \item[] Question: For each theoretical result, does the paper provide the full set of assumptions and a complete (and correct) proof?
    \item[] Answer: \answerNA{} 
    \item[] Justification: N/A, because our work focuses solely on empirical evaluation to measure the effectiveness of the proposed framework.
    \item[] Guidelines:
    \begin{itemize}
        \item The answer NA means that the paper does not include theoretical results. 
        \item All the theorems, formulas, and proofs in the paper should be numbered and cross-referenced.
        \item All assumptions should be clearly stated or referenced in the statement of any theorems.
        \item The proofs can either appear in the main paper or the supplemental material, but if they appear in the supplemental material, the authors are encouraged to provide a short proof sketch to provide intuition. 
        \item Inversely, any informal proof provided in the core of the paper should be complemented by formal proofs provided in appendix or supplemental material.
        \item Theorems and Lemmas that the proof relies upon should be properly referenced. 
    \end{itemize}

    \item {\bf Experimental result reproducibility}
    \item[] Question: Does the paper fully disclose all the information needed to reproduce the main experimental results of the paper to the extent that it affects the main claims and/or conclusions of the paper (regardless of whether the code and data are provided or not)?
    \item[] Answer: \answerYes{} 
    \item[] Justification: Yes, this information can be found in \cref{sec:experiment} and in Appendix.
    \item[] Guidelines:
    \begin{itemize}
        \item The answer NA means that the paper does not include experiments.
        \item If the paper includes experiments, a No answer to this question will not be perceived well by the reviewers: Making the paper reproducible is important, regardless of whether the code and data are provided or not.
        \item If the contribution is a dataset and/or model, the authors should describe the steps taken to make their results reproducible or verifiable. 
        \item Depending on the contribution, reproducibility can be accomplished in various ways. For example, if the contribution is a novel architecture, describing the architecture fully might suffice, or if the contribution is a specific model and empirical evaluation, it may be necessary to either make it possible for others to replicate the model with the same dataset, or provide access to the model. In general. releasing code and data is often one good way to accomplish this, but reproducibility can also be provided via detailed instructions for how to replicate the results, access to a hosted model (e.g., in the case of a large language model), releasing of a model checkpoint, or other means that are appropriate to the research performed.
        \item While NeurIPS does not require releasing code, the conference does require all submissions to provide some reasonable avenue for reproducibility, which may depend on the nature of the contribution. For example
        \begin{enumerate}
            \item If the contribution is primarily a new algorithm, the paper should make it clear how to reproduce that algorithm.
            \item If the contribution is primarily a new model architecture, the paper should describe the architecture clearly and fully.
            \item If the contribution is a new model (e.g., a large language model), then there should either be a way to access this model for reproducing the results or a way to reproduce the model (e.g., with an open-source dataset or instructions for how to construct the dataset).
            \item We recognize that reproducibility may be tricky in some cases, in which case authors are welcome to describe the particular way they provide for reproducibility. In the case of closed-source models, it may be that access to the model is limited in some way (e.g., to registered users), but it should be possible for other researchers to have some path to reproducing or verifying the results.
        \end{enumerate}
    \end{itemize}

\item {\bf Open access to data and code}
    \item[] Question: Does the paper provide open access to the data and code, with sufficient instructions to faithfully reproduce the main experimental results, as described in supplemental material?
    \item[] Answer: \answerYes{} 
    \item[] Justification: Yes, we will release our code upon acceptance.
    \item[] Guidelines:
    \begin{itemize}
        \item The answer NA means that paper does not include experiments requiring code.
        \item Please see the NeurIPS code and data submission guidelines (\url{https://nips.cc/public/guides/CodeSubmissionPolicy}) for more details.
        \item While we encourage the release of code and data, we understand that this might not be possible, so “No” is an acceptable answer. Papers cannot be rejected simply for not including code, unless this is central to the contribution (e.g., for a new open-source benchmark).
        \item The instructions should contain the exact command and environment needed to run to reproduce the results. See the NeurIPS code and data submission guidelines (\url{https://nips.cc/public/guides/CodeSubmissionPolicy}) for more details.
        \item The authors should provide instructions on data access and preparation, including how to access the raw data, preprocessed data, intermediate data, and generated data, etc.
        \item The authors should provide scripts to reproduce all experimental results for the new proposed method and baselines. If only a subset of experiments are reproducible, they should state which ones are omitted from the script and why.
        \item At submission time, to preserve anonymity, the authors should release anonymized versions (if applicable).
        \item Providing as much information as possible in supplemental material (appended to the paper) is recommended, but including URLs to data and code is permitted.
    \end{itemize}

\item {\bf Experimental setting/details}
    \item[] Question: Does the paper specify all the training and test details (e.g., data splits, hyperparameters, how they were chosen, type of optimizer, etc.) necessary to understand the results?
    \item[] Answer: \answerYes{} 
    \item[] Justification: Yes, we provided them in \cref{sec:experiment} and in Appendix.
    \item[] Guidelines:
    \begin{itemize}
        \item The answer NA means that the paper does not include experiments.
        \item The experimental setting should be presented in the core of the paper to a level of detail that is necessary to appreciate the results and make sense of them.
        \item The full details can be provided either with the code, in appendix, or as supplemental material.
    \end{itemize}

\item {\bf Experiment statistical significance}
    \item[] Question: Does the paper report error bars suitably and correctly defined or other appropriate information about the statistical significance of the experiments?
    \item[] Answer: \answerNo{} 
    \item[] Justification: No, because the evaluation process takes hours on a GPU and is very expensive to compute with multiple seeds. Importantly, we conducted evaluation for several benchmarks in \cref{sec:experiment} so running with different seeds are not feasible in our scope.
    \item[] Guidelines:
    \begin{itemize}
        \item The answer NA means that the paper does not include experiments.
        \item The authors should answer "Yes" if the results are accompanied by error bars, confidence intervals, or statistical significance tests, at least for the experiments that support the main claims of the paper.
        \item The factors of variability that the error bars are capturing should be clearly stated (for example, train/test split, initialization, random drawing of some parameter, or overall run with given experimental conditions).
        \item The method for calculating the error bars should be explained (closed form formula, call to a library function, bootstrap, etc.)
        \item The assumptions made should be given (e.g., Normally distributed errors).
        \item It should be clear whether the error bar is the standard deviation or the standard error of the mean.
        \item It is OK to report 1-sigma error bars, but one should state it. The authors should preferably report a 2-sigma error bar than state that they have a 96\% CI, if the hypothesis of Normality of errors is not verified.
        \item For asymmetric distributions, the authors should be careful not to show in tables or figures symmetric error bars that would yield results that are out of range (e.g. negative error rates).
        \item If error bars are reported in tables or plots, The authors should explain in the text how they were calculated and reference the corresponding figures or tables in the text.
    \end{itemize}

\item {\bf Experiments compute resources}
    \item[] Question: For each experiment, does the paper provide sufficient information on the computer resources (type of compute workers, memory, time of execution) needed to reproduce the experiments?
    \item[] Answer: \answerYes{} 
    \item[] Justification: Yes, we provided them in the Appendix.
    \item[] Guidelines:
    \begin{itemize}
        \item The answer NA means that the paper does not include experiments.
        \item The paper should indicate the type of compute workers CPU or GPU, internal cluster, or cloud provider, including relevant memory and storage.
        \item The paper should provide the amount of compute required for each of the individual experimental runs as well as estimate the total compute. 
        \item The paper should disclose whether the full research project required more compute than the experiments reported in the paper (e.g., preliminary or failed experiments that didn't make it into the paper). 
    \end{itemize}
    
\item {\bf Code of ethics}
    \item[] Question: Does the research conducted in the paper conform, in every respect, with the NeurIPS Code of Ethics \url{https://neurips.cc/public/EthicsGuidelines}?
    \item[] Answer: \answerYes{} 
    \item[] Justification: Yes, we totally confirm to it.
    \item[] Guidelines:
    \begin{itemize}
        \item The answer NA means that the authors have not reviewed the NeurIPS Code of Ethics.
        \item If the authors answer No, they should explain the special circumstances that require a deviation from the Code of Ethics.
        \item The authors should make sure to preserve anonymity (e.g., if there is a special consideration due to laws or regulations in their jurisdiction).
    \end{itemize}

\item {\bf Broader impacts}
    \item[] Question: Does the paper discuss both potential positive societal impacts and negative societal impacts of the work performed?
    \item[] Answer: \answerYes{} 
    \item[] Justification: We discussed the societal imparts in \cref{sec:conclusion}.
    \item[] Guidelines:
    \begin{itemize}
        \item The answer NA means that there is no societal impact of the work performed.
        \item If the authors answer NA or No, they should explain why their work has no societal impact or why the paper does not address societal impact.
        \item Examples of negative societal impacts include potential malicious or unintended uses (e.g., disinformation, generating fake profiles, surveillance), fairness considerations (e.g., deployment of technologies that could make decisions that unfairly impact specific groups), privacy considerations, and security considerations.
        \item The conference expects that many papers will be foundational research and not tied to particular applications, let alone deployments. However, if there is a direct path to any negative applications, the authors should point it out. For example, it is legitimate to point out that an improvement in the quality of generative models could be used to generate deepfakes for disinformation. On the other hand, it is not needed to point out that a generic algorithm for optimizing neural networks could enable people to train models that generate Deepfakes faster.
        \item The authors should consider possible harms that could arise when the technology is being used as intended and functioning correctly, harms that could arise when the technology is being used as intended but gives incorrect results, and harms following from (intentional or unintentional) misuse of the technology.
        \item If there are negative societal impacts, the authors could also discuss possible mitigation strategies (e.g., gated release of models, providing defenses in addition to attacks, mechanisms for monitoring misuse, mechanisms to monitor how a system learns from feedback over time, improving the efficiency and accessibility of ML).
    \end{itemize}
    
\item {\bf Safeguards}
    \item[] Question: Does the paper describe safeguards that have been put in place for responsible release of data or models that have a high risk for misuse (e.g., pretrained language models, image generators, or scraped datasets)?
    \item[] Answer: \answerYes{} 
    \item[] Justification: Yes, our code will be released with Licences.
    \item[] Guidelines:
    \begin{itemize}
        \item The answer NA means that the paper poses no such risks.
        \item Released models that have a high risk for misuse or dual-use should be released with necessary safeguards to allow for controlled use of the model, for example by requiring that users adhere to usage guidelines or restrictions to access the model or implementing safety filters. 
        \item Datasets that have been scraped from the Internet could pose safety risks. The authors should describe how they avoided releasing unsafe images.
        \item We recognize that providing effective safeguards is challenging, and many papers do not require this, but we encourage authors to take this into account and make a best faith effort.
    \end{itemize}

\item {\bf Licenses for existing assets}
    \item[] Question: Are the creators or original owners of assets (e.g., code, data, models), used in the paper, properly credited and are the license and terms of use explicitly mentioned and properly respected?
    \item[] Answer: \answerYes{} 
    \item[] Justification: Yes, we have properly cited the code, data and models used.
    \item[] Guidelines:
    \begin{itemize}
        \item The answer NA means that the paper does not use existing assets.
        \item The authors should cite the original paper that produced the code package or dataset.
        \item The authors should state which version of the asset is used and, if possible, include a URL.
        \item The name of the license (e.g., CC-BY 4.0) should be included for each asset.
        \item For scraped data from a particular source (e.g., website), the copyright and terms of service of that source should be provided.
        \item If assets are released, the license, copyright information, and terms of use in the package should be provided. For popular datasets, \url{paperswithcode.com/datasets} has curated licenses for some datasets. Their licensing guide can help determine the license of a dataset.
        \item For existing datasets that are re-packaged, both the original license and the license of the derived asset (if it has changed) should be provided.
        \item If this information is not available online, the authors are encouraged to reach out to the asset's creators.
    \end{itemize}

\item {\bf New assets}
    \item[] Question: Are new assets introduced in the paper well documented and is the documentation provided alongside the assets?
    \item[] Answer: \answerNA{} 
    \item[] Justification: At the time of submission, no new assets are being released. We plan to release the associated code publicly if the paper is accepted.
    \item[] Guidelines:
    \begin{itemize}
        \item The answer NA means that the paper does not release new assets.
        \item Researchers should communicate the details of the dataset/code/model as part of their submissions via structured templates. This includes details about training, license, limitations, etc. 
        \item The paper should discuss whether and how consent was obtained from people whose asset is used.
        \item At submission time, remember to anonymize your assets (if applicable). You can either create an anonymized URL or include an anonymized zip file.
    \end{itemize}

\item {\bf Crowdsourcing and research with human subjects}
    \item[] Question: For crowdsourcing experiments and research with human subjects, does the paper include the full text of instructions given to participants and screenshots, if applicable, as well as details about compensation (if any)? 
    \item[] Answer: \answerNA{} 
    \item[] Justification: The paper does not involve crowdsourcing nor research with human subjects.
    \item[] Guidelines:
    \begin{itemize}
        \item The answer NA means that the paper does not involve crowdsourcing nor research with human subjects.
        \item Including this information in the supplemental material is fine, but if the main contribution of the paper involves human subjects, then as much detail as possible should be included in the main paper. 
        \item According to the NeurIPS Code of Ethics, workers involved in data collection, curation, or other labor should be paid at least the minimum wage in the country of the data collector. 
    \end{itemize}

\item {\bf Institutional review board (IRB) approvals or equivalent for research with human subjects}
    \item[] Question: Does the paper describe potential risks incurred by study participants, whether such risks were disclosed to the subjects, and whether Institutional Review Board (IRB) approvals (or an equivalent approval/review based on the requirements of your country or institution) were obtained?
    \item[] Answer: \answerNA{} 
    \item[] Justification: The paper does not involve crowdsourcing nor research with human subjects.
    \item[] Guidelines:
    \begin{itemize}
        \item The answer NA means that the paper does not involve crowdsourcing nor research with human subjects.
        \item Depending on the country in which research is conducted, IRB approval (or equivalent) may be required for any human subjects research. If you obtained IRB approval, you should clearly state this in the paper. 
        \item We recognize that the procedures for this may vary significantly between institutions and locations, and we expect authors to adhere to the NeurIPS Code of Ethics and the guidelines for their institution. 
        \item For initial submissions, do not include any information that would break anonymity (if applicable), such as the institution conducting the review.
    \end{itemize}

\item {\bf Declaration of LLM usage}
    \item[] Question: Does the paper describe the usage of LLMs if it is an important, original, or non-standard component of the core methods in this research? Note that if the LLM is used only for writing, editing, or formatting purposes and does not impact the core methodology, scientific rigorousness, or originality of the research, declaration is not required.
    \item[] Answer: \answerNA{} 
    \item[] Justification: The paper does not involve LLM in the core of this method.
    \item[] Guidelines:
    \begin{itemize}
        \item The answer NA means that the core method development in this research does not involve LLMs as any important, original, or non-standard components.
        \item Please refer to our LLM policy (\url{https://neurips.cc/Conferences/2025/LLM}) for what should or should not be described.
    \end{itemize}

\end{enumerate}

%% file: tex/appendix.tex
\section{Proof of \cref{prop:accum_cfg}}
Given a shortcut model $\stheta$ and its corresponding classifier-free guided output $\gtheta^w$ with guidance scale $w$. To analyze the generation process, we define a sequence of intermediate points $\{\bm{x}'_{\frac{i}{N}}\}_{i=0}^N$. This sequence is constructed by recursively applying $N$ consecutive shortcut steps, each with the smallest step size $d=1/N$. Note that $N$ is chosen to be a power of 2. The process begins from the initial noise $\xzero$ at $t=0$, conditioned on $c$. The starting point of the sequence is thus defined as $\bm{x}'_0 = \xzero$, and subsequent intermediate points are generated as follow:
\begin{equation}
    \bm{x}'_{\frac{i+1}{N}} = \bm{x}'_\frac{i}{N} + \stheta\left(\bm{x}'_\frac{i}{N}, \frac{i}{N}, c, d\right)d, \quad \text{for } i=0, \dots, N-1.
\end{equation}

Ideally, we assume that the model is perfectly trained, or equivalently, that the loss in \cref{eq:shortcut} is minimized to zero. Under this assumption, we have:
\begin{align}
    \stheta(\xt', t, c, 2d)&=\frac{1}{2}[\gtheta^w(\xt', t, c, d) + \gtheta^w(\bm{x}'_{t+d}, t+d, c, d)], \label{eq:ideal_cond_1} \\
    \stheta(\xt', t, \varnothingcond, 2d)&=\frac{1}{2}[\stheta(\xt', t, \varnothingcond, d)+\stheta(\bm{x}'_{t+d}, t+d, \varnothingcond, d)]. \label{eq:ideal_cond_2}
\end{align}
In the following derivation, we omit the time $t$ in the network notation for simplicity, unless otherwise specified. First, we will prove by induction that:
\begin{equation}
    \gtheta^{w}(\xt', c, 2^jd) = \frac{1}{2^j} \sum_{i=0}^{2^j-1} \gtheta^{w^{j+1}} \left( \bm{x}'_{t+id}, c, d \right).
    \label{eq:inductive_g}
\end{equation}
For the base case $j=0$, we have $\gtheta^{w}(\xt', c, d) = \gtheta^{w} (\xt', c, d)$, which satisfies \cref{eq:inductive_g}. Assuming that \cref{eq:inductive_g} holds for $j = k$, we now show that it also holds for $j = k+1$. Using \cref{eq:ideal_cond_1} and \cref{eq:ideal_cond_2}, we obtain:
\begin{align*}
\stheta(\xt', c, 2^{k+1}d) &= \frac{1}{2} [\gtheta^w(\xt', c, 2^kd) + \gtheta^w(\bm{x}'_{t+2^kd}, c, 2^kd)] \\
&= \frac{1}{2^{k+1}} \left[\sum_{i=0}^{2^{k}-1} \gtheta^{w^{k+1}} \left(\bm{x}'_{t+id}, c, d \right)+\sum_{i=0}^{2^{k}-1} \gtheta^{w^{k+1}} \left(\bm{x}'_{t+2^kd+id}, c, d \right)\right] \\
&= \frac{1}{2^{k+1}} \sum_{i=0}^{2^{k+1}-1} \gtheta^{w^{k+1}} \left(\bm{x}'_{t+id}, c, d \right), \\
\stheta(\xt', \varnothingcond, 2^{k+1}d) &= \frac{1}{2} \left[\stheta(\xt', \varnothingcond, 2^kd)+\stheta(\bm{x}'_{t+2^kd}, \varnothingcond, 2^kd)\right] \\
&= \frac{1}{2^{k+1}} \sum_{i=0}^{2^{k+1}-1}\stheta(\bm{x}'_{t+id}, \varnothingcond, d).
\end{align*}
Therefore, we have:
\begin{align*}
&\gtheta^{w}(\xt', c, 2^{k+1}d) \\
&= w\stheta(\xt', c, 2^{k+1}d) + (1-w)\stheta(\xt', \varnothingcond, 2^{k+1}d) \\
&= \frac{1}{2^{k+1}} \sum_{i=0}^{2^{k+1}-1}[w\gtheta^{w^{k+1}} \left(\bm{x}'_{t+id}, c, d \right)+ (1-w)\stheta(\bm{x}'_{t+id}, \varnothingcond, d)]  \\
&= \frac{1}{2^{k+1}} \sum_{i=0}^{2^{k+1}-1}[w(w^{k+1}\stheta(\bm{x}'_{t+id}, c, d)+(1-w^{k+1})\stheta(\bm{x}'_{t+id}, \varnothingcond, d))+ (1-w)\stheta(\bm{x}'_{t+id}, \varnothingcond, d)] \\
&= \frac{1}{2^{k+1}} \sum_{i=0}^{2^{k+1}-1}[w^{k+2}\stheta(\bm{x}'_{t+id}, c, d)+(1-w^{k+2})\stheta(\bm{x}'_{t+id}, \varnothingcond, d)] \\
&= \frac{1}{2^{k+1}} \sum_{i=0}^{2^{k+1}-1} \gtheta^{w^{k+2}} \left(\bm{x}'_{t+id}, c, d \right). \\
\end{align*}
This satisfies \cref{eq:inductive_g} for $j=k+1$. By induction, we have \cref{eq:inductive_g} holds for all $j$. Based on this result, we can rewrite the output of the shortcut model for a single large step of size $Nd = 1$, denoted by $\stheta(\xzero, c, Nd)$, as follows:
\begin{align*}
    \stheta(\xzero, c, Nd) &= \frac{1}{2}\left[\gtheta^w\left(\xzero, c, \frac{N}{2}d\right) + \gtheta^w\left(\bm{x}'_{\frac{1}{2}}, c, \frac{N}{2}d\right)\right] \\
    &=\frac{1}{N}\left[\sum_{i=0}^{N/2-1} \gtheta^{w^{\log_2(N/2)+1}} \left(\bm{x}'_{\frac{i}{N}}, c, d\right) + \sum_{i=0}^{N/2-1} \gtheta^{w^{\log_2(N/2)+1}} \left(\bm{x}'_{\frac{1}{2}+\frac{i}{N}}, c, d\right)\right]\\
    &=\frac{1}{N}\sum_{i=0}^{N-1}\gtheta^{w^{\log_2(N)}}\left(\bm{x}'_{\frac{i}{N}}, c, d\right),
\end{align*}
which completes the proof.

\section{Multi-Level Wavelet Function}

\textbf{Multi-Level Wavelet Function.} \cref{alg:mutil_code} provides pseudo-code for our proposed multi-level wavelet objective described in \cref{sec:method-wavelet}.

\begin{figure}[h]
    \vspace{-0.5cm}
    \begin{minipage}{\textwidth}
        \begin{algorithm}[H]
            \caption{Multi-level Wavelet Function}
            \label{alg:mutil_code} 
            \definecolor{codeblue}{rgb}{0.25,0.5,0.5}
            \definecolor{codegreen}{rgb}{0,0.6,0}
            \definecolor{codekw}{RGB}{207,33,46}
            \definecolor{highlighter}{rgb}{1,1,0.6}
            \lstset{
              backgroundcolor=\color{white},
              basicstyle=\fontsize{7.5pt}{7.5pt}\ttfamily\selectfont,
              columns=fullflexible,
              breaklines=true,
              captionpos=b,
              commentstyle=\fontsize{7.5pt}{7.5pt}\color{codegreen},
              keywordstyle=\fontsize{7.5pt}{7.5pt}\color{codekw},
              xleftmargin=-0.2cm, 
              xrightmargin=0.01cm, 
              escapechar={|}, 
              frame=None,
              numbers=none,
            }
            \begin{lstlisting}[language=python]
    class MultiLevelWaveletLoss:
        def __init__(self):
            self.dwt = DWT_2D("haar")  # Discrete Wavelet Transform (DWT)
            self.diff_func = MSELoss()  # Distance function

        def concatenated_dwt(self, x):
            xll, xlh, xhl, xhh = self.dwt(x)  # Decompose into 4 wavelet sub-bands using DWT
            details = torch.cat([xll, xlh, xhl, xhh], dim=1)  # Concatenate the sub-bands
            return details
            
        def __call__(self, pred, target, num_levels):
            total_loss = diff_func(pred, target).mean()  # Calculate low-level loss in original output space

            # Recursively calculate loss on wavelet sub-bands for each level
            pred_curr, target_curr = pred, target
            for current_level in range(num_levels):
                # Derive predicted and target sub-bands using outputs from previous level
                pred_bands = self.concatenated_dwt(pred_curr)
                target_bands = self.concatenated_dwt(target_curr)

                # Calculate loss on current level
                total_loss += diff_func(pred_bands, target_bands).mean()
                pred_curr, target_curr = pred_bands, target_bands

            # Taking average from all levels
            loss = total_loss / (num_levels + 1)
            return loss
            \end{lstlisting}
        \end{algorithm}
    \end{minipage}
\end{figure}

\begin{figure}
    \centering
    \vspace{-23pt}
    \includegraphics[width=0.95\textwidth]{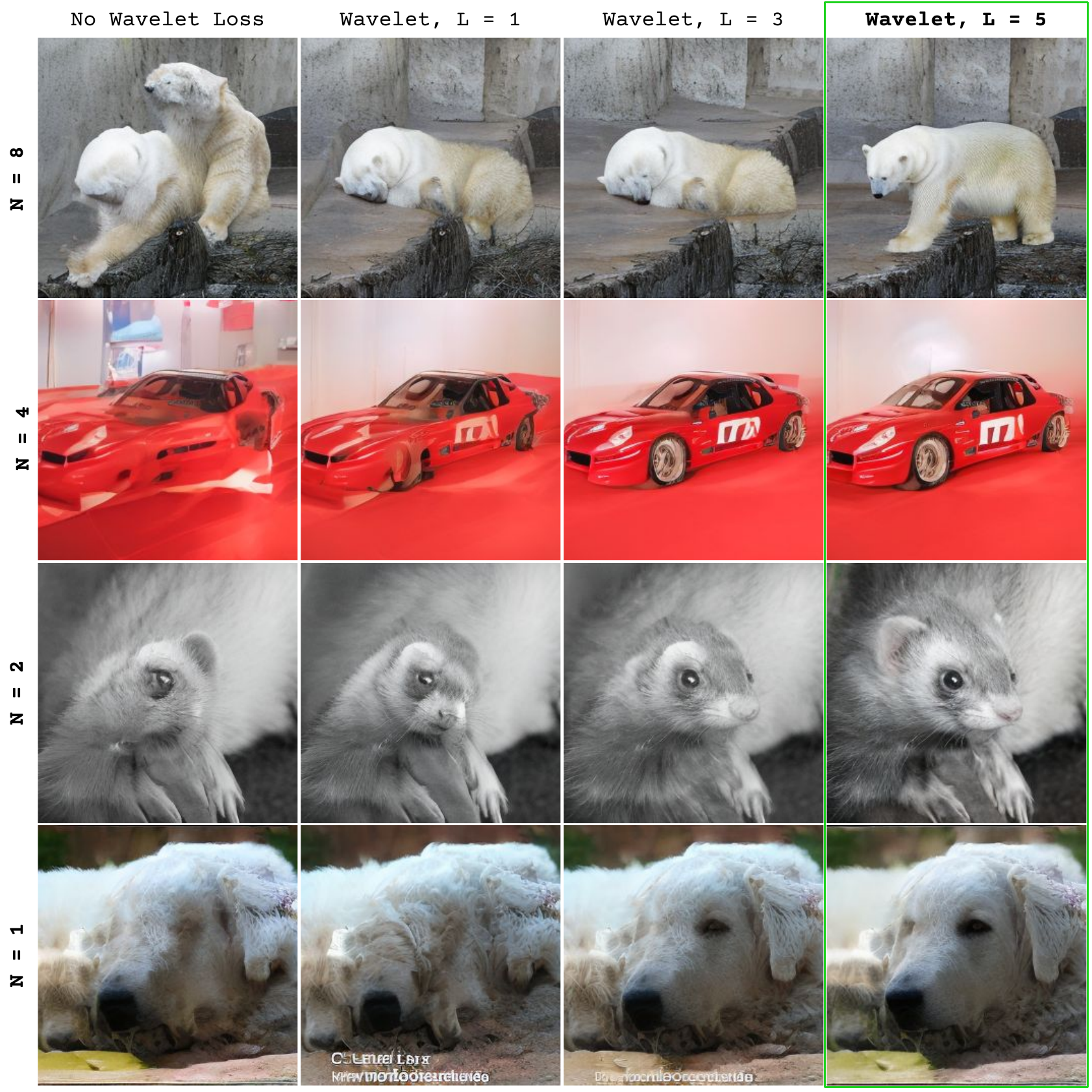}
    \vspace{3pt}
    \caption{Effect of applying different numbers of wavelet loss layers ($L$) on image quality under one and few-step inference. The traditional low-level objective (leftmost column) results in noticeable degradation and artifacts. In contrast, using multi-level wavelet loss, especially with $L=5$ (green box, right), produces high-quality, consistent images across all inference settings.}
    \vspace{-13pt}
    \label{fig:freq}
\end{figure}

\textbf{Results.} \cref{fig:freq} presents qualitative comparisons illustrating the impact of different levels in our multi-level wavelet function versus traditional low-level loss. Incorporating more wavelet levels yields finer details and fewer artifacts, especially in one- and few-step generation.

\section{Injecting Conditional Inputs into Network}
We explore two primary strategies for incorporating conditional information including CFG scale $w$, current sample $\xt$, time $t$, condition $c$, and the target step size $d$ into our network. The first strategy, similar to U-ViT \cite{bao2023all}, encodes each condition as an individual token appended to the input sequence of noisy image patch tokens. The second strategy employs AdaLN-Zero blocks \cite{peebles2023scalable} to modulate the network with each condition; these modulations are then aggregated through addition. We find that the AdaLN-Zero approach yields comparable performance to the U-ViT method but without the drawback of increased input sequence length. Given this efficiency benefit, we adopt AdaLN-Zero for injecting CFG information into our network.
\clearpage
\vspace{-7pt}
\section{Accelerating training speed}
\vspace{-7pt}
\input{table/top_up}
To mitigate the high computational cost of shortcut models, we explore resuming their training from existing a flow matching weights \cite{yu2025repa}. \cref{tab:top_up} shows that while resuming offers limited improvement for the original shortcut model, our improved shortcut models (iSM) demonstrate considerable gains. Specifically, when incorporating a series of proposed methods, iSM achieves significantly better FID scores. Models are trained for 250k steps for all evaluations.
 \vspace{-10pt}
\section{Experiment Settings}
\subsection{Training \& Parameterization Setting}
For experiments on ImageNet, images are preprocessed to 256x256 resolution, following the protocol of ADM \cite{dhariwal2021diffusion}. Our models operate in the latent space derived from the sd-vae-ft-mse autoencoder \citep{stabilityai2023sdvaeftmse}. These latents are subsequently normalized channel-wise using means of $[0.86488,\ -0.27787343,\ 0.21616915,\ 0.3738409]$ and standard deviations of $[4.85503674,\ 5.31922414,\ 3.93725398,\ 3.9870003]$. Finally, the normalized latents are scaled by a factor of 0.5, targeting an approximate standard deviation of 0.5 for the network input. Detailed configurations for all experiments are provided in \cref{tab:expsetup}.

\input{table/settings}

\subsection{Evaluation Details} We follow the ADM~\citep{dhariwal2021diffusion} evaluation setup, using the same reference batches from their official implementation,\footnote{\url{https://github.com/openai/guided-diffusion/tree/main/evaluations}} and compute FID~\citep{fid} over 50K generated images.

\subsection{Baselines} Below, we summarize the baseline methods used in our evaluation.
\begin{itemize}
\item \textbf{ADM}~\citep{dhariwal2021diffusion} improves U-Net-based diffusion architectures and introduces classifier-guided sampling to balance sample quality and diversity.
\item \textbf{CDM}~\citep{ho2022cascaded} proposes cascaded diffusion models, which generate images in a coarse-to-fine manner, similar to ProgressiveGAN~\citep{karras2018progressive}.
\item \textbf{Simple diffusion}~\citep{hoogeboom2023simple} leverages a diffusion model for high-resolution images by simplifying the noise schedule and model architectures.
\item \textbf{LDM}~\citep{rombach2022high} introduces the concept of operating diffusion models in a compressed latent space, enhancing efficiency.
\item \textbf{U-DiT}~\citep{tian2024u} proposes a series of U-shaped DiTs based on self-attention with downsampled tokens.
\item \textbf{U-ViT}~\citep{bao2023all} adapts Vision Transformers for latent diffusion by incorporating U-Net-like long skip connections.
\item \textbf{DiT}~\citep{peebles2023scalable} pioneers the use of a pure transformer architecture as the backbone for diffusion models, featuring AdaIN-zero modules.
\item \textbf{SiT}~\citep{ma2024sit} investigates how to improve DiT training by transitioning from discrete diffusion to continuous flow-based modeling.
\item \textbf{REPA}~\citep{yu2025repa} accelerates the training of DiT/SiT models by regularizing network representations to align with features from pretrained visual encoders.
\item \textbf{FlowDCN}~\citep{flowdcn} offers a fully convolutional architecture for generative modeling with linear time and memory complexity, enabling efficient high-resolution image synthesis.
\item \textbf{iCT}~\citep{song2023improved} introduces several techniques to enhance the training of CMs~\citep{song2023consistency}, including a lognormal noise schedule, Pseudo-Huber loss functions, and a scheduler for total discretization steps during training.
\item \textbf{SM}~\citep{shortcut} establishes a framework for one-step generation by combining flow matching with a self-consistency objective.
\item \textbf{IMM}~\citep{imm} few-step generative models by inductively matching all moments of bootstrapped samples derived from stochastic interpolants using Maximum Mean Discrepancy (MMD), aiming for stable convergence.
\end{itemize}

\section{Related Works}
\subsection{Diffusion, Consistency \& Flow Matching models} Generative modeling has seen significant advancements with methods like diffusion models, consistency models, and flow matching. Diffusion models~\citep{sohl2015deep,song2020score,ho2020denoising,kingma2021variational} are a generative framework that synthesizes data by gradually transforming random noise into outputs via a stochastic denoising process. To introduce efficient generation, consistency models~\citep{song2023consistency,song2023improved,lu2024simplifying} aim for efficient generation by learning a mapping from any point on a solution trajectory directly to the data origin. Meanwhile, flow matching~\citep{lipman2022flow,liu2022flow} proposes learning generative models by defining a target vector field between noise and data and training a neural network to approximate this field. Both these methods have demonstrated strong performance when scaled to text-to-image~\citep{rombach2022high,saharia2022photorealistic,podell2023sdxl,chen2023pixart,esser2024scaling} and text-to-video~\citep{ho2022imagen,blattmann2023stable,sora} applications.

\subsection{End-to-end training for one-to-many step generative models} 
The quest for efficient, high-quality few-step generation has explored several avenues. While early one-step models leveraged GANs~\citep{goodfellow2020generative,karras2020analyzing,brock2018large} and MMD~\citep{li2015generative,li2017mmd}, the inherent instability and complexity of adversarial training limit their scalability. Consistency Models (CMs)~\citep{song2023consistency, song2023improved, lu2024simplifying} address the instability of adversarial training without needing synthetic datasets. Unlike distillation, CMs can be trained from scratch via consistency training (CT), independent of pre-trained diffusion models. However, few-step generation with CMs usually involves discrete-time variants. These variants require meticulous timestep scheduling and are susceptible to irreducible bias accumulation due to the inherent ambiguity of discretization. Inductive Moment Matching (IMM) \cite{imm} further advanced stable few-step training using moment matching. Shortcut Models (SMs) \cite{shortcut} introduced step-size conditioning with self-bootstrapping. However, original SMs (Section \ref{sec:method-guidance}) suffer from inflexible guidance and guidance accumulation artifacts, which our work directly addresses.




\section{More Qualitative Results}
\begin{figure}
    \centering
    \vspace{-25pt}
    \includegraphics[width=0.95\textwidth]{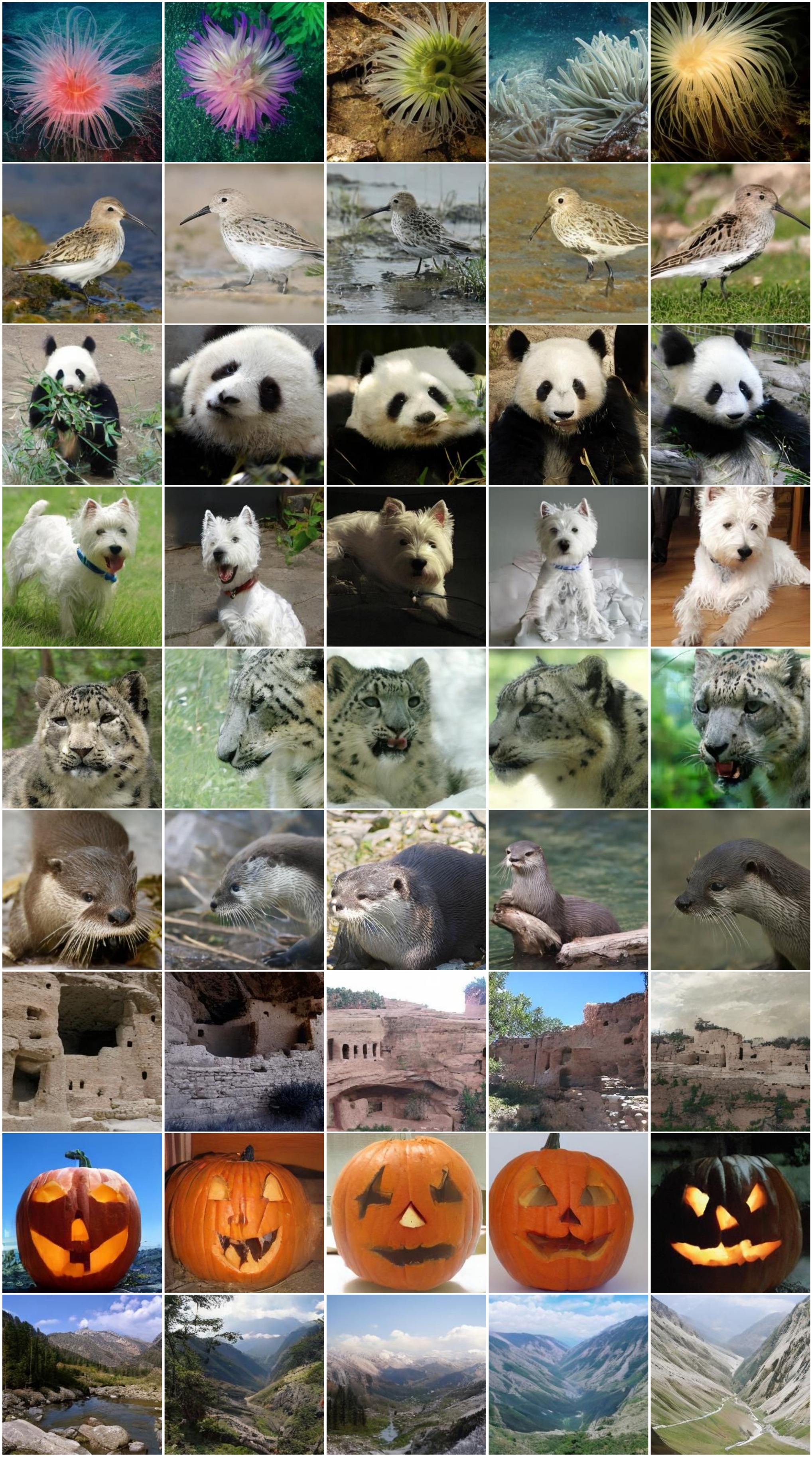}
    \caption{Uncurated samples on ImageNet-$256\times256$ with 1-step sampling.}
    
\end{figure}

\begin{figure}
    \centering
    \vspace{-25pt}
    \includegraphics[width=0.95\textwidth]{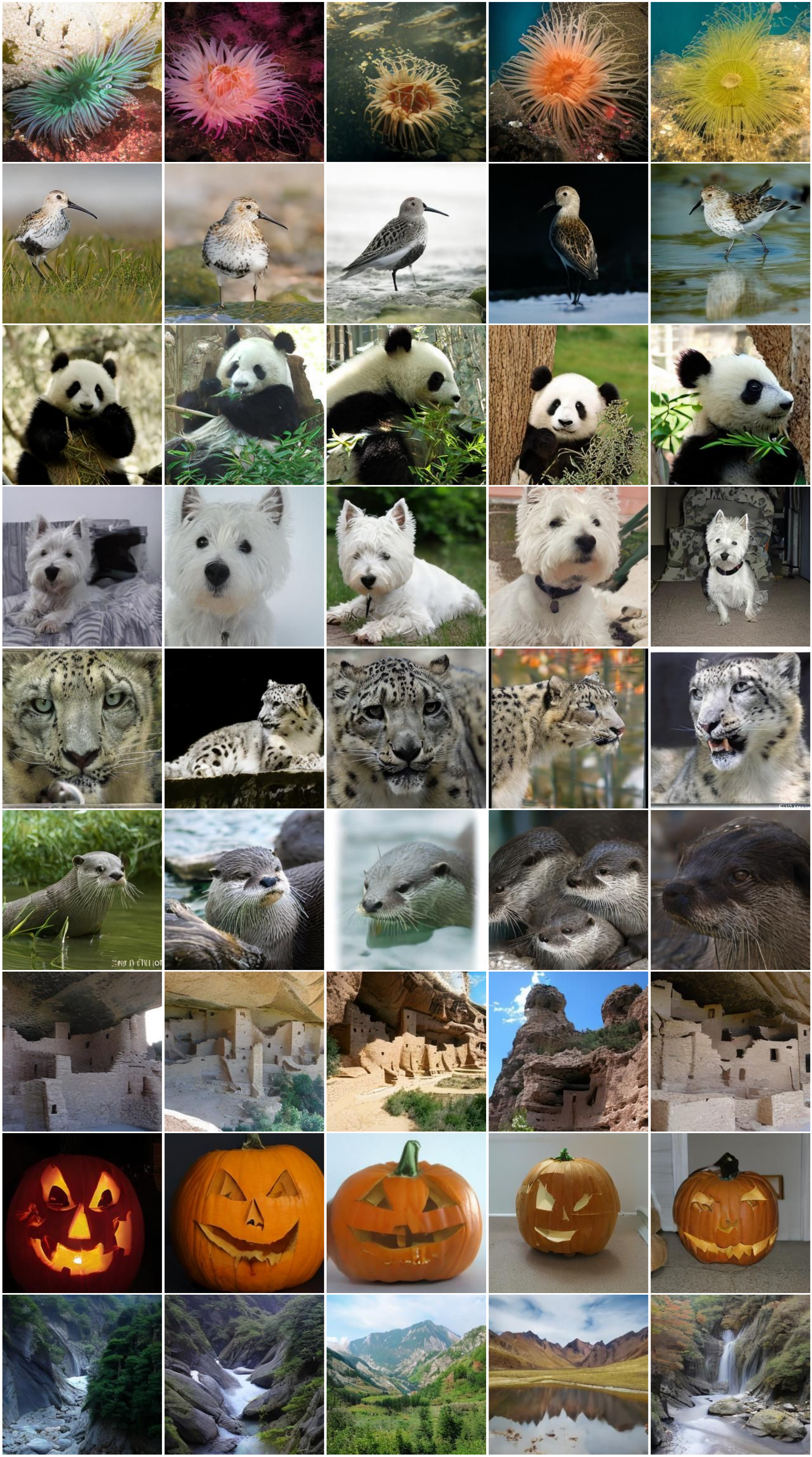}
    \caption{Uncurated samples on ImageNet-$256\times256$ with 2-step sampling.}
    
\end{figure}

\begin{figure}
    \centering
    \vspace{-25pt}
    \includegraphics[width=0.95\textwidth]{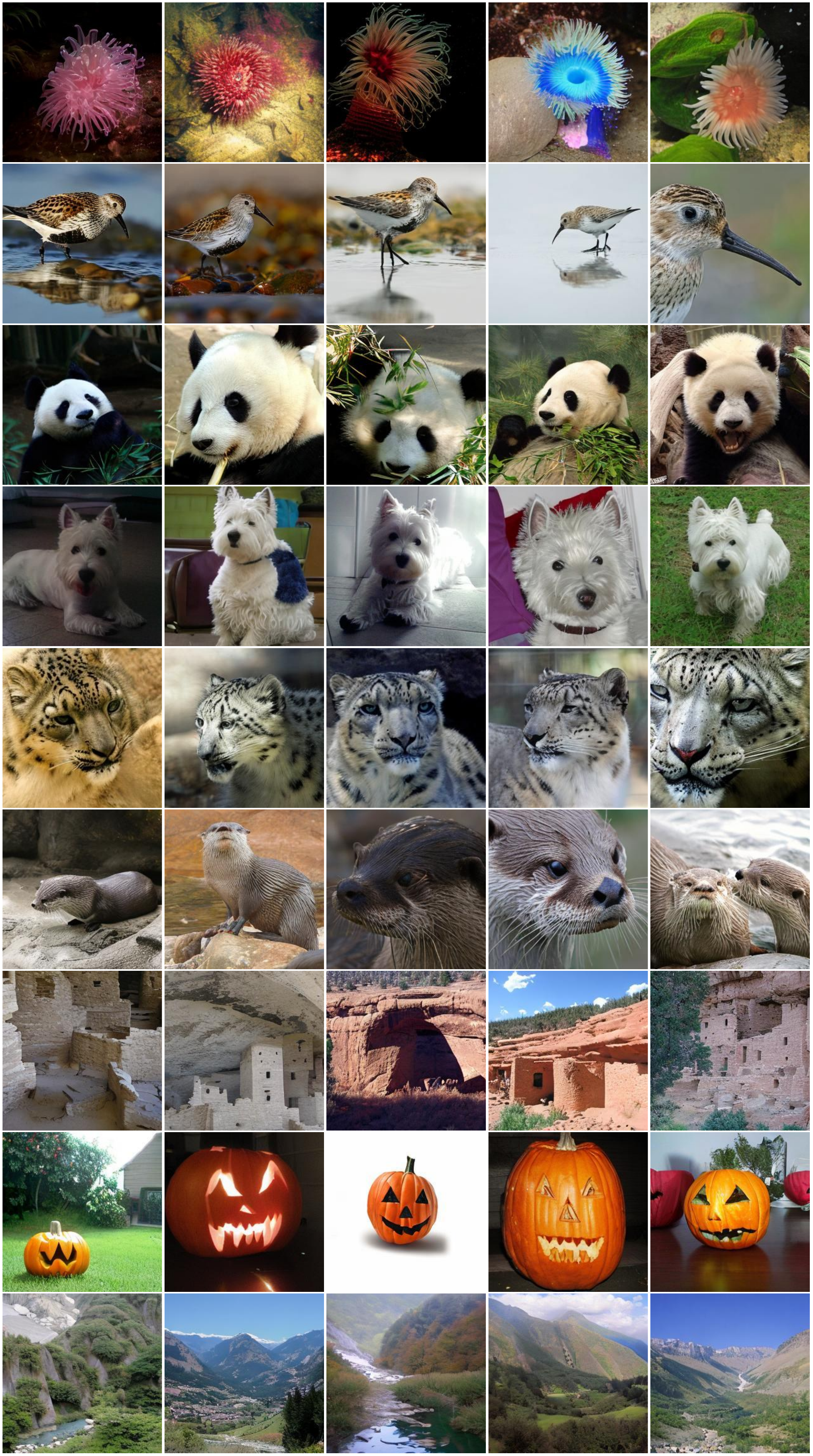}
    \caption{Uncurated samples on ImageNet-$256\times256$ with 4-step sampling.}
    
\end{figure}

\begin{figure}
    \centering
    \vspace{-25pt}
    \includegraphics[width=0.95\textwidth]{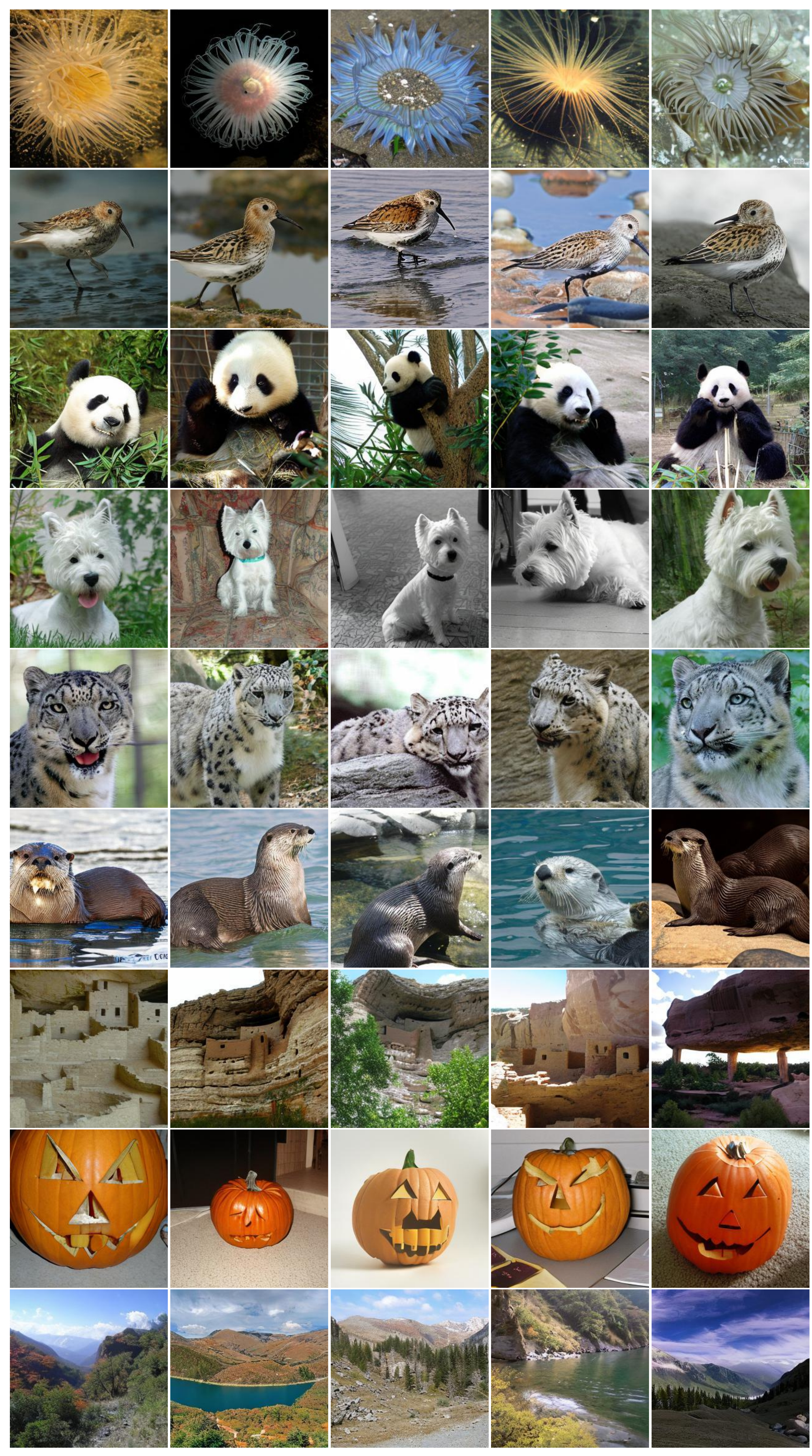}
    \caption{Uncurated samples on ImageNet-$256\times256$ with 8-step sampling.}
\end{figure}

\begin{figure}
    \centering
    \vspace{-25pt}
    \includegraphics[width=0.95\textwidth]{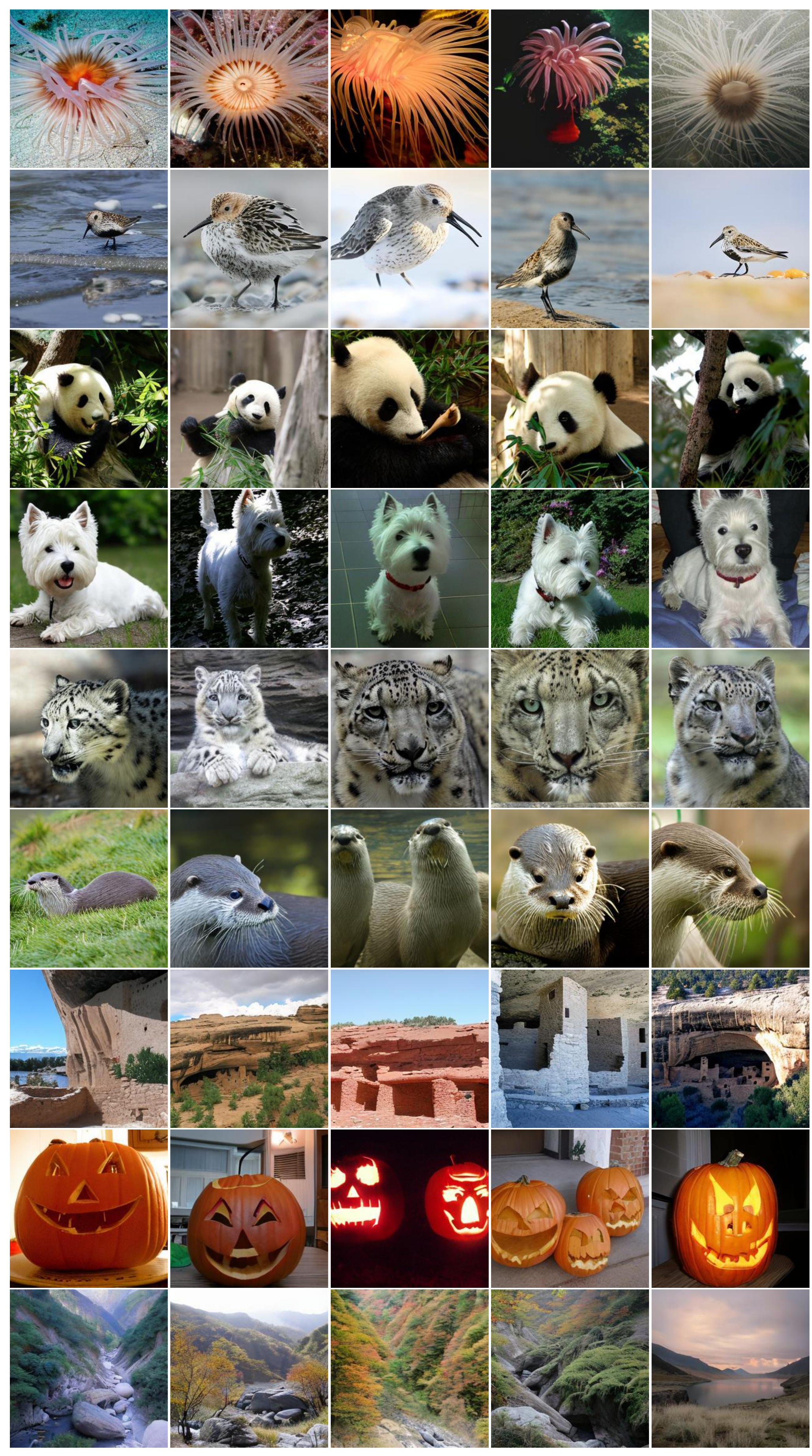}
    \caption{Uncurated samples on ImageNet-$256\times256$ with 128-step sampling.}
\end{figure}

%% file: table/top_up.tex
\begin{wraptable}{r}{7cm}
    \centering
    \scriptsize
    \captionsetup{font=small}
    
        \vspace{-12pt}
        \caption{Performance of Improved Shortcut Models}
        \vspace{5pt}

        \begin{tabular}{lcr} 
            \toprule
            \textbf{Method} & \textbf{FID$_{\one}$ $\downarrow$} & \textbf{FID$_{\four}$ $\downarrow$} \\
            \midrule
            \rowcolor{gray!15} Shortcut Models \cite{shortcut} & 21.38 & 13.46 \\
            \midrule
            \multicolumn{3}{@{}c@{}}{\textbf{\textit{Improved Shortcut Models (iSM)}}} \\ 
            \midrule
            + Intrinsic Guidance & 9.62 & 3.17 \\
            + Interval Guidance in Training & 8.49 & 2.81 \\
            + Multi-level Wavelet Function & 8.12 & 2.64 \\
            + Scaling Optimal Transport  & 7.97 & 2.23 \\
            + Twin EMA & 6.56 & 2.16 \\
            \bottomrule
        \end{tabular}

    \label{tab:top_up}
\end{wraptable}

%% file: table/settings.tex
\begin{table}[t!]
    \centering
    \small
    \scriptsize
    \caption{Experimental settings on ImageNet $256\times256$.}
    \vspace{3.0pt}
    \resizebox{0.7\textwidth}{!}{
    \begin{tabular}{lcccccc}
    \toprule
        \multicolumn{3}{l}{\textbf{Parameterization Setting}}\\
        \midrule
        Architecture &  SiT-XL \\  
        GFlops & 118.64  \\  
        Params (M) & 675 \\
        Flow Trajectory & OT-FM \\
        Input dim. & $32\ttimes 32\ttimes 4$ \\
        Num. layers & $24$ \\
        Hidden dim. & $1024$ \\
        Num. heads & $16$ \\
        $\alpha_t$ & $1-t$ \\
        $\sigma_t$ & $t$ \\
        $w_t$ & $\sigma_t$ \\
        Training objective & v-prediction \\
        
        \midrule
        \multicolumn{3}{l}{\textbf{Training Setting}}\\
        \midrule
        Training iteration & 800K \\
        Dropout & $0$ \\  
        Optimizer & AdamW \\ 
        AdamW $\beta_1$ & $0.9$ \\ 
        AdamW $\beta_2$ & $0.999$ \\ 
        AdamW $\epsilon$ & $10^{-8}$  \\ 
        Learning Rate & $0.0001$ \\ 
        Weight Decay & $0$ \\ 
        Batch Size & $256$ \\
        Label Dropout & $0.1$ \\
        \midrule
        \multicolumn{3}{l}{\textbf{Methods}}\\
        \midrule
        CFG Scale $w_{\wmax}$ & $3.5$ \\
        Interval $t_{\interval}$ & $0.3$ \\
        Wavelet Levels $L$ & $5$ \\
        OT Scale $K$ & $32$ \\
        Ratio of Empirical to Self-consistency Targets & $0.25$ \\
        EMA Parameters Used For Self-consistency Targets? & True \\
        EMA Target Rate $\theta^{-}_{\target}$ & $0.95$ \\
        EMA Parameters Used For Evaluation? & True \\
        EMA Inference Rate $\theta^{-}_{\infer}$ & $0.9999$ \\
        \bottomrule
    \end{tabular}
    }
    \label{tab:expsetup}
\end{table}